\documentclass[journal]{IEEEtran}
\usepackage{graphicx}
\usepackage[ruled]{algorithm2e}
\usepackage{array}
\usepackage{subfig}
\usepackage{booktabs}
\usepackage{url}
\usepackage{xr-hyper}
\usepackage{hyperref}
\usepackage{multirow}
\usepackage{longtable}
\usepackage[table,xcdraw,dvipsnames]{xcolor}
\usepackage{tabularx}
\usepackage{amsmath}
\usepackage{amsfonts}
\usepackage{amssymb}
\usepackage{bbm}
\usepackage{capt-of}
\usepackage{nccmath}

\makeatletter
\newcommand*{\addFileDependency}[1]{
  \typeout{(#1)}
  \@addtofilelist{#1}
  \IfFileExists{#1}{}{\typeout{No file #1.}}
}
\makeatother

\newcommand*{\myexternaldocument}[1]{%
    \externaldocument{#1}%
    \addFileDependency{#1.tex}%
    \addFileDependency{#1.aux}%
}

\myexternaldocument{./supp_info}

\ifCLASSOPTIONcompsoc
  \usepackage[nocompress]{cite}
\else
  \usepackage{cite}
\fi
\ifCLASSINFOpdf
\else
\fi
\begin{document}

\begin{titlepage}

An Evaluation of Anomaly Detection and Diagnosis in Multivariate Time Series by Astha Garg, Wenyu Zhang, Jules Samaran, Ramasamy Savitha, Chuan-Sheng Foo.
DOI: 10.1109/TNNLS.2021.3105827

© 2021 IEEE. Personal use of this material is permitted. Permission from IEEE must be obtained for all other uses, in any current or future media, including reprinting/republishing this material for advertising or promotional purposes, creating new collective works, for resale or redistribution to servers or lists, or reuse of any copyrighted component of this work in other works.

Supplemental information for this manuscript is available at \url{https://github.com/astha-chem/mvts-ano-eval/tree/main/supp_info}
\end{titlepage}

%
\title{An Evaluation of Anomaly Detection and Diagnosis in Multivariate Time Series}
%
%
%
%

\author{Astha~Garg, 
        Wenyu~Zhang, 
        Jules~Samaran,
        Ramasamy~Savitha,~\IEEEmembership{Senior~Member,~IEEE,}
        and Chuan-Sheng~Foo
\IEEEcompsocitemizethanks{\IEEEcompsocthanksitem A. Garg is at Chord X Pte. Ltd. and was at Institute of Infocomm Research (I2R), Agency for Science, Technology and Research (A*STAR), Singapore when the work was conducted. E-mail: astha.iitb@gmail.com
\IEEEcompsocthanksitem W. Zhang, C.S. Foo and Ramasamy S. are at I2R, A*STAR, Singapore. E-mail:  \{ramasamysa, foocs\}@i2r.a-star.edu.sg
\IEEEcompsocthanksitem J. Samaran is with Mines Paristech, PSL Research University, France.}
}

%
%

\markboth{Accepted at IEEE TNNLS Special Issue on Deep Learning for Anomaly Detection Aug 2021}%
{Garg \MakeLowercase{\textit{et al.}}: Submitted Version}
%



\IEEEtitleabstractindextext{%
\begin{abstract}
Several techniques for multivariate time series anomaly detection have been proposed recently, but a systematic comparison on a common set of datasets and metrics is lacking. This paper presents a systematic and comprehensive evaluation of unsupervised and semi-supervised deep-learning based methods for anomaly detection and diagnosis on multivariate time series data from cyberphysical systems. Unlike previous works, we vary the \textit{model} and post-processing of model errors, i.e. the \textit{scoring functions} independently of each other, through a grid of 10 models and 4 scoring functions, comparing these variants to state of the art methods. In time-series anomaly detection, detecting anomalous events is more important than detecting individual anomalous time-points. Through experiments, we find that the existing evaluation metrics either do not take events into account, or cannot distinguish between a good detector and trivial detectors, such as a random or an all-positive detector. We propose a new metric to overcome these drawbacks, namely, the composite F-score ($Fc_1$), for evaluating time-series anomaly detection. 

Our study highlights that dynamic scoring functions work much better than static ones for multivariate time series anomaly detection, and the choice of scoring functions often matters more than the choice of the underlying model. We also find that a simple, channel-wise model - the Univariate Fully-Connected Auto-Encoder, with the dynamic Gaussian scoring function emerges as a winning candidate for both anomaly detection and diagnosis, beating state of the art algorithms.
\end{abstract}

\begin{IEEEkeywords}
Multivariate Time Series, Anomaly Detection, Evaluation, Anomaly Diagnosis, Deep Learning, Metrics
\end{IEEEkeywords}}

\maketitle

\IEEEdisplaynontitleabstractindextext

%
\IEEEpeerreviewmaketitle
\section{Introduction}
\label{sec:introduction}
Modern cyberphysical systems (CPS), such as those encountered in manufacturing, aircraft and servers, involve sophisticated equipment that records multivariate time-series (MVTS) data from 10s, 100s, or even thousands of sensors. The MVTS need to be continuously monitored to ensure smooth operation and prevent expensive failures. We focus on two key monitoring tasks on MVTS in this work. First, identifying time-points during operation where sensors indicate deviation from normal behavior, which we refer to as the streaming time-series \textit{anomaly detection} task. Second, pointing out the specific channel(s)\footnote{A channel refers to the time-series corresponding to a particular variable.} that deviate from normal behavior, which would aid the operator in verifying the anomaly, finding its root-cause and taking corrective action; we refer to this as the \textit{anomaly diagnosis} task. In addition, while the data under healthy operation from sensor monitoring is abundant, there is typically a lack of labeled data for anomalous operation. We thus focus on the \emph{semi-supervised} setting, where the training data consists only of data from healthy operation \cite{chalapathy2019deepano,pang2021}, and the \emph{unsupervised} anomaly detection setting, where the training data is mostly healthy but may contain a small number of unknown anomalies \cite{chalapathy2019deepano}. 

Amongst recent works on semi-supervised and unsupervised MVTS anomaly detection, deep learning based methods have featured prominently   \cite{li2019mad,zhou2019beatgan, hundman2018detecting,park2018lstmvae,malhotra2016lstm,audibert2020usad,zhang2019deep,su2019omni,tariq2019mixpca,tuor2017onlinethreat,zheng2019ocan,ren2019microsoft}, and several of these are state-of-the-art. 
In this work, we aim to characterize the key design choices behind these deep learning methods and their effect on performance for anomaly detection and diagnosis tasks, to better understand the reasons for their success. Specifically, we observe that a majority of deep MVTS anomaly detection methods \cite{li2019mad,zhang2019deep,zhou2019beatgan,hundman2018detecting,park2018lstmvae, malhotra2016lstm,audibert2020usad} work by learning feature representations of normality\cite{pang2021}, under the assumption that a model trained to reconstruct or forecast patterns in healthy data would have a high error on anomalous test data\cite{hawkins2002outlier, chandola2009anomaly}. For this class of 
methods, the channel-wise model errors must be transformed and combined across channels using a \textit{scoring function} (defined formally in Section \ref{sec:framework}) to obtain a single anomaly score per time-point. The score must then be thresholded for anomaly detection. Anomaly diagnosis may be carried out by ranking the channel-wise anomaly scores (or errors) for the duration of the anomaly, before aggregation, and returning the top ranked channels for anomaly diagnosis. Our observations raise several interesting questions:

\begin{itemize}
\item What is an appropriate scoring function to use for anomaly detection and diagnosis? 

\item How important is the choice of scoring functions compared to the choice of model for good performance?

\item How well do existing algorithms perform on anomaly detection and diagnosis on data from real CPS systems? 

\end{itemize}
\vspace{-2mm}
\subsection{Contributions}
\label{sec:contributions}
To address the questions identified above, we carry out the most comprehensive evaluation of deep semi-supervised algorithms for MVTS anomaly detection and diagnosis to date, using real-world, publicly available CPS datasets. We show that deep anomaly detection methods that work by learning feature representations of normality can be put into a modular framework consisting of 3 parts -- a model, a scoring function and a thresholding function. Using this framework, we cross 10 distinct models against 4 scoring functions to investigate the effect of independent choices of the model and the scoring function. We compare these combinations to other recently proposed end-to-end algorithms. In total, we evaluate 45 and 29 unique end-to-end algorithms on 7 and 4 datasets for MVTS anomaly detection and diagnosis respectively. The code will be released at \url{https://github.com/astha-chem/mvts-ano-eval}. We make several interesting discoveries: 
\begin{itemize}
    \item We find that \textit{the choice of an appropriate scoring function can (a) boost the anomaly detection performance of existing methods, and (b) might matter more than the choice of the underlying model}. In particular, dynamic scoring functions, i.e. scoring functions that adapt to variations in the test set, outperform static scoring functions overall. To the best of our knowledge, dynamic scoring functions have not been investigated for deep anomaly detection for MVTS before. 
    \item Surprisingly, we find that the \textit{Univariate Fully-Connected AutoEncoder (UAE) -- a simple model, when used with dynamic scoring  outperforms all other algorithms overall on both anomaly detection and diagnosis.} UAE consists of independent channel-wise fully-connected auto-encoder models. This is a straightforward approach, but has not been comprehensively evaluated before for MVTS anomaly detection.
    \item In order to identify an appropriate metric for our evaluation, we compared existing F-score metrics for time series anomaly detection in terms of their robustness and ability to reward the detection of anomalous events. Significantly, we find that a popular metric, point-adjusted $F_1$ score\cite{xu2018donut,su2019omni,audibert2020usad} gives a close to perfect score of 0.96 to an anomaly detector that predicts anomalies randomly on one of the datasets. Thus, we find that the existing evaluation metrics either do not take events into account ($F_1$ score), or are not robust (point-adjusted $F_1$ score), and \textit{we propose a new simple, yet robust metric for evaluating anomalous event detection -- the \textit{composite} $F_1$ score, $Fc_1$. }
\end{itemize} 

\vspace{-1mm}
\section{Related Work}
\label{sec:related_work}
Recent surveys on general anomaly detection\cite{chandola2009anomaly}, deep-learning based anomaly detection \cite{chalapathy2019deepano,pang2021} and unsupervised time-series anomaly detection \cite{blazquez2020reviewtsano} review techniques relevant to unsupervised and semi-supervised MVTS anomaly detection. Time-series anomaly detection techniques may be categorized based on their detection technique, namely, shallow model-based \cite{ahmad2017numenta,Hochenbaum2017Twitter,laptev2015generic,zhang2019gpmultivar}, deep-learning model based \cite{li2019mad,zhang2019deep,zhou2019beatgan,su2019omni, hundman2018detecting,park2018lstmvae,audibert2020usad,ren2019microsoft,tariq2019mixpca,kieu2019ensembleaed,xu2018donut}, pattern-based \cite{feremans2019pattern,yeh2016matrix}, distance based \cite{keogh2005discord,wang2018cpumultivar} and non-parametric \cite{siffer2017evt}. Of these, deep-learning based techniques have received significant attention for MVTS anomaly detection owing to (a) their ability to scale to high dimensions and model complex patterns in various domains, compared to straightforward statistical approaches such as out of limits approaches \cite{venkatasubramanian2003processpart1, qin2012industrial, hundman2018detecting}, (b) fast inference and applicability to streaming time-series typical in CPS unlike many distance-based and pattern-based techniques that are not applicable to streaming time-series, as they require both training and test data during inference\cite{keogh2005discord,wang2018cpumultivar,feremans2019pattern,yeh2016matrix}, and (c) the ability to localize anomalous time points within sequences, unlike techniques \cite{feremans2019pattern,yeh2016matrix,zhao2019pyod} that work at the coarser level to detect anomalous sub-sequences.
Anomaly diagnosis has been approached primarily from a supervised classification perspective \cite{de2008damadics-hmm}. In the unsupervised context, four studies \cite{su2019omni,tariq2019mixpca,tuor2017onlinethreat,zhang2019deep} mention that ranking of scores or errors can be used to diagnose the cause of anomalies but only \cite{su2019omni} shows experimental results on an open dataset.

\textbf{Choice of algorithms for evaluation}: We evaluate semi-supervised and unsupervised deep anomaly detection techniques for MVTS, applicable to our problem setup of streaming time-series, localization of anomalies within sequences and no anomalies in the training set. In terms of the categorization proposed by \cite{pang2021} for deep anomaly detection we include a range of techniques, summarized in Figure \ref{fig:taxonomy} in the supplemental information (SI). We include (a) algorithms that work by \textit{generic normality feature learning}, using auto-encoders -- LSTM-ED\cite{malhotra2016lstm},  LSTM-VAE \cite{vaelstm2019}, MSCRED \cite{zhang2019deep}; using Generative Adversarial Networks (GAN) -- BeatGAN \cite{zhou2019beatgan}; and using predictability modeling -- NASA LSTM \cite{hundman2018detecting}, (b) technique that \textit{learns anomaly-measure dependent features} -- DAGMM\cite{zong2018deep} (c) \textit{end-to-end anomaly scoring techniques} -- OCAN \cite{zheng2019ocan} and OmniAnomaly\cite{su2019omni}. We include representative shallow techniques -- Principal Component Analysis (PCA)\cite{li2014model} and One-Class Support Vector Machine (OC-SVM)\cite{manevitz2002ocsvm}. In addition, we compare published results for OmniAnomaly, USAD\cite{audibert2020usad} and MAD-GAN\cite{li2019mad} with ours in section \ref{sec:compare_published} in the SI. We did not evaluate some recently proposed algorithms as their source-code is proprietary and is not straightforward to implement \cite{tariq2019mixpca,ren2019microsoft}. 

\section{Problem setup}
\label{sec:problem_setup}
\textbf{Anomaly detection}: We are given a training MVTS, $\textbf{X}^{train}\in \mathbb{R}^{n_1 \times m}$ with $n_1$ regularly sampled time-points and $m$ channels. $\textbf{X}^{train}$ is known or assumed to contain no anomalies. The task is to predict whether an anomaly occurred at each time point $t$ in the test time-series $\textbf{X}^{test} \in       \mathbb{R}^{n_2 \times m}$ with $n_2$ time-points and $1 \leq t \leq n_2$. When making a prediction for time $t$, we assume that the test time-series has only been observed until
time $t$ to simulate a streaming scenario \cite{ahmad2017numenta}; we also assume that $\textbf{X}^{train}$ is not available at test time.

\noindent \textbf{Anomaly diagnosis}: We consider the anomaly diagnosis task independently; it is also referred to as anomaly interpretation \cite{su2019omni}. Given $\textbf{X}^{train}$ and $\textbf{X}^{test}$ as above, as well as the start and end times of each anomalous event, we want to predict the specific channel(s) that deviate from normal behavior. In this paper, we refer to these as \textit{causes}. Note that we do not imply these to be the root causes, but rather that these are the causes due to which the algorithm flagged an anomaly. 
\section{Datasets}
\label{sec:datasets}

\begin{table*}[h]
\setlength{\tabcolsep}{4pt}
\centering
\caption{Summary of dataset characteristics used in this paper. Averaging is done over entities.}
\label{tab:datasets}
\resizebox{\textwidth}{!}{%
\begin{tabular}{@{}lp{0.15\textwidth}llp{0.08\textwidth}p{0.07\textwidth}p{0.07\textwidth}p{0.08\textwidth}p{0.07\textwidth}p{0.06\textwidth}lll@{}}
\toprule
 Name & Domain & Entities & Channels, $m$ & Average train length & Average test length & Average \% anomalies & Average num events & Event time (mins) & Time step (s) & $l_w$ & $l_s$\\ \midrule
SWaT\cite{goh2016swat}  & Water treatment   & 1  & 51  & 473400  & 414569 & 4.65\% & 35    & 1.7-28.1  & 1  & 100 & 10\\
WADI\cite{ahmed2017wadi}  & Water distribution   & 1  & 123 & 1209601 & 172801 & 5.76\%  & 14    & 1.5-29   & 1  & 30 & 10\\
DMDS\cite{damadics} & Sugar manufacturing & 1  & 32  & 507600  & 217802 & 1.48\%  & 17    & 0.2-10.3 & 1  & 100 & 10\\
SKAB\cite{skab} & Water circulation    & 1 & 8 & 9401 & 35600 & 36.70\% & 34 & 2.4-9.8 & 1 & 100 & 1\\
MSL\cite{hundman2018detecting}  & Spacecraft    & 27 & 55  & 2160    & 2731   & 12.02\% & 1.33  & 11-1141  & 60 & 100 & 1\\
SMAP\cite{hundman2018detecting} & Spacecraft    & 55 & 25  & 2556    & 8071   & 12.40\% & 1.26  & 31-4218  & 60 & 100 & 1\\
SMD\cite{su2019omni} & Server monitoring & 28 & 38  & 25300   & 25301  & 4.21\%  & 11.68 & 2-3160 & 60 & 100 & 1\\ \bottomrule
\end{tabular}
}
\end{table*}
We use 7 publicly available MVTS datasets from real-world CPS (Table \ref{tab:datasets}), characterized by a regular sampling rate, periodicity in several channels and strong correlations across time and channels (e.g. SI Fig. \ref{fig:sample_ts}). The training sets are known or assumed to be anomaly-free. A few channels in these datasets stay constant in the training set but they may help to detect anomalies in the test set and are not discarded. To model each time-series, we break it into overlapping windows of length $l_w$ and a step size, $l_s$ such that $1 \leq l_s \leq l_w$ (Table \ref{tab:datasets}). We consider three \textit{multi-entity} datasets (MSL, SMAP, SMD), where each entity is a different physical unit of the same type, having the same dimensionality. Similar to \cite{hundman2018detecting} and \cite{su2019omni}, we train a separate model for each entity in the multi-entity datasets. The other four datasets are \textit{single-entity} datasets. 

The anomalies were induced knowingly by physically compromising the operation in a steady-state system, for datasets other than SMAP and MSL. The ground truth of the root cause is known for SWaT, WADI, DMDS and SMD datasets. For SMAP and MSL, the anomalies are expert-labeled manually based on past reports of actual spacecraft operation\cite{hundman2018detecting}. Unlike the other datasets, each entity in MSL and SMAP consists of only 1 sensor, while all the other channels are one-hot-encoded commands given to that entity. For MSL and SMAP, we use all channels as input to the models, but use the model error of only the sensor channel for anomaly detection. The datasets are summarized in Table \ref{tab:datasets} and additional details including train-test splits are provided in the SI section \ref{sec:supp_ds_info}. 

Broadly, there are two types of anomalies in MVTS datasets - \textit{temporal anomalies}, where one or more channels deviate from their normal behavior when compared to their respective history (e.g. malfunctioning of a monitored part in the system), and \textit{cross-channel anomalies}, where each channel individually looks normal with respect to its history, but its relationship to other channels is abnormal. Based on author descriptions of anomalies for SWaT\cite{goh2016swat}, WADI\cite{ahmed2017wadi}, DMDS\cite{damadics}, MSL \cite{hundman2018detecting} and SMAP \cite{hundman2018detecting} datasets, and our own visualizations (e.g. Fig \ref{Fig:demo}), the datasets we test primarily contain \textit{temporal anomalies}. While we expect temporal anomalies to be more common in steady state CPS than strictly cross-channel anomalies, datasets showing multiple operational states such as powerplant operation in  \cite{zhang2019gpmultivar} may be more likely to have cross-channel anomalies; we did not find open datasets where cross-channel anomalies are known to be present. 
\section{Modular framework}
\label{sec:framework}
\begin{figure}[h]
    \centering
  \includegraphics[width=0.8\linewidth]{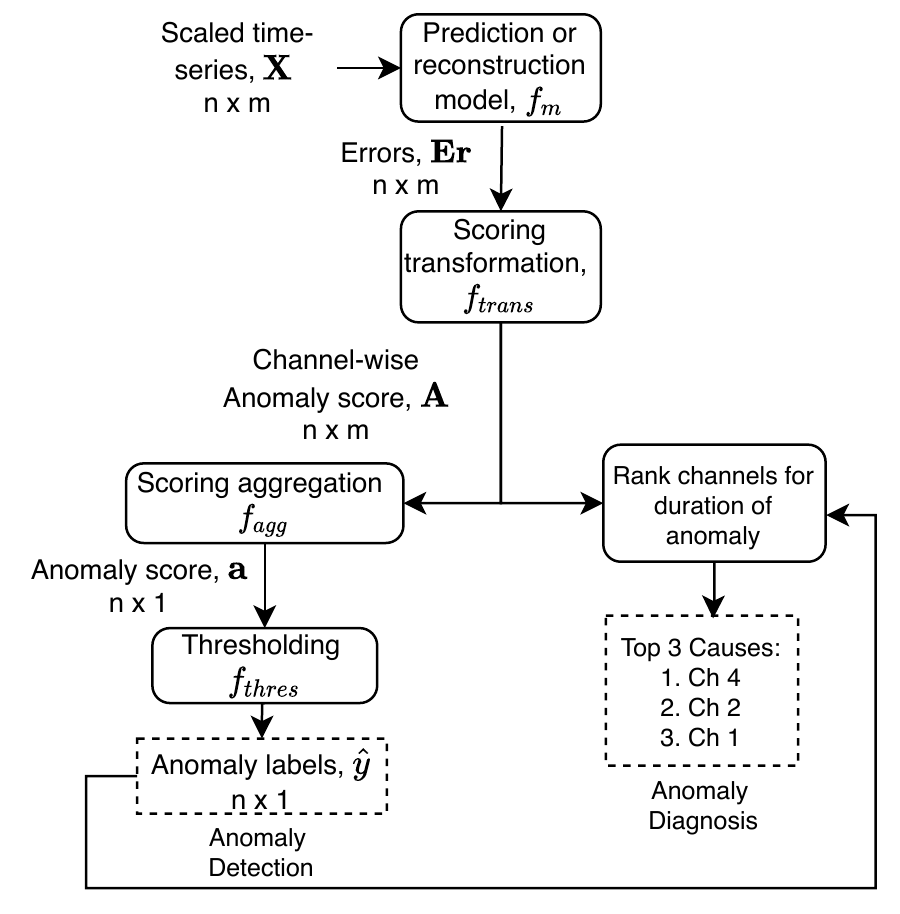}
\caption{Framework for anomaly detection and diagnosis.}
\label{fig:overall-schematic}       
\end{figure}

A majority of deep anomaly detection approaches for MVTS are based on training a model to reconstruct or predict healthy time-series \cite{li2019mad, zhang2019deep,zhou2019beatgan,hundman2018detecting,park2018lstmvae,malhotra2016lstm,audibert2020usad,su2019omni}. We generalize these model-based approaches into a modular framework for anomaly detection and diagnosis consisting of 3 parts, or modules - a reconstruction or prediction model, a scoring function and a thresholding function (Fig. \ref{fig:overall-schematic}). 

The first module is a \textit{reconstruction or prediction model}, $f_m$, that takes as input a sub-sequence, and outputs a lossy reconstruction of the input or a prediction of the next (or several) time-point(s). $f_m$ is used to calculate an error, $\mathbf{Er}_t^i$, where the superscript $i$ denotes channels and the subscript $t$ denotes the $t^{th}$ time-point. $f_m$ is not limited to neural network models and could be, for example, PCA\cite{li2014model}, ARIMA\cite{yu2016arima} or HMM\cite{ahmad2017numenta} models. 

The second module, \textit{scoring function} combines a multi-channel score such as errors $\mathbf{Er}$ (or latent representation as in DAGMM \cite{zong2018deep}) into a single anomaly score $\mathbf{a}_t$ per time-point. The scoring function usually has two distinct but related parts - a transformation function $f_{trans}$ and an aggregation function $f_{agg}$. First, $f_{trans}$ transforms the errors into the channel-wise anomaly score, $\mathbf{A}_t^i$, which can be used for anomaly diagnosis to identify which channel has the highest anomaly score for the duration of the anomaly. Then, $f_{agg}$ aggregates the scores across channels resulting in the anomaly score, $\mathbf{a}_t$ for each time-point.   

The final module, the \textit{thresholding function}, $f_{thres}$ thresholds $\mathbf{a}_t$ to obtain a binary label $\hat{\mathbf{y}}_t$ to classify each point as anomalous or healthy. Formally, for $\mathbf{X} = \mathbf{X_{test}}$, 
\begin{subequations}
\label{eqn:framework}
\begin{align}
\mathbf{Er}(\mathbf{X}) = \mathbf{X} - f_m(\mathbf{X}) \quad &\text{;} \quad \mathbf{A}(\mathbf{X}) = f_{trans}(\mathbf{Er}(\mathbf{X}))\\
\mathbf{a}(\mathbf{X}) = f_{agg}(\mathbf{A}(\mathbf{X})) \quad &\text{;} \quad \hat{\mathbf{y}} = f_{thres}(\mathbf{a}(\mathbf{X}))
\end{align}
\end{subequations}

\begin{table}[h]
\centering
\caption{Algorithms evaluated in this work}
\label{tab:algos_evals}
\resizebox{\linewidth}{!}{%
\begin{tabular}{p{0.65\linewidth}p{0.35\linewidth}}
Model + Scoring function & Threshold\\

\toprule

\textbf{Models crossed with scoring functions} & \\ 
Models: Raw Signal, PCA\cite{li2014model}, UAE, FC AE, LSTM AE\cite{malhotra2016lstm}, TCN\cite{bai2018tcnvsrnn} AE, LSTM VAE\cite{park2018lstmvae}, BeatGAN\cite{zhou2019beatgan}, MSCRED\cite{zhang2019deep}, NASA LSTM\cite{hundman2018detecting}. 
\newline Scoring: Error, Gauss-S, Gauss-D, Gauss-D-K
& best-F-score, top-k, tail p    \\ \midrule
\textbf{Models with pre-defined scoring} & \\
DAGMM\cite{zong2018deep},  OmniAnomaly\cite{su2019omni}, OCAN\cite{zheng2019ocan}  & best-F-score, top-k, tail p \\
NASA   LSTM NPT\cite{hundman2018detecting}         & Non-parametric threshold\\ 
OC-SVM\cite{manevitz2002ocsvm}     & Thresholding at 0.5 \\\bottomrule
\end{tabular}
}
\end{table} 

\section{Algorithms}
\label{sec:algorithms}
First we discuss the models for which we vary scoring functions in section \ref{sec: independent-scoring}, followed by models with pre-defined scoring functions in section \ref{sec: predefined-scoring}. We also relate these models to the categorization proposed by \cite{pang2021}, shown in SI Fig. \ref{fig:taxonomy}. We discuss scoring functions and thresholding functions separately in sections \ref{sec: scoring}  and \ref{sec:thresholding} respectively. All the algorithms evaluated are summarized in Table \ref{tab:algos_evals}. 

\subsection{Models where scoring functions are varied}
\label{sec: independent-scoring}
\subsubsection{Shallow models} We test \textbf{Raw Signal} as a trivial ``model" that reconstructs any signal to 0, so that the error is the same as the normalized signals. We also test \textbf{PCA} for lossy reconstruction \cite{li2014model} by retaining only a subset of principal components (PC) that explain 90\% of the variance. 

\subsubsection{Generic normality feature learning models}
An auto-encoder (AE) \cite{rumelhart1986autoencoder} is an unsupervised deep neural network that is trained to reconstruct the input through a compressed latent representation, using an encoder and a decoder. The AE is the basis of many deep-learning based models for MVTS anomaly detection \cite{audibert2020usad,zhang2019deep,malhotra2016lstm}, thus we include various architectures for this. In the \textbf{Univariate Fully-Connected AE (UAE)} model, we train a separate AE for each channel. \textbf{Fully-Connected AE (FC AE)} takes data concatenated across channels as input, and hence can capture relationships between channels. \textbf{Long Short Term Memory AE (LSTM AE)} is based on the LSTM Encoder-Decoder model by \cite{malhotra2016lstm}. While \cite{malhotra2016lstm} used only the first principal component (PC) of the MVTS as input to LSTM-ED, we instead set 90\% explained variance for PCA to minimize information loss. \textbf{Temporal Convolutional Network AE (TCN AE)} is based on the TCN model proposed by \cite{bai2018tcnvsrnn}, where we stack TCN residual blocks for the encoder, and we replace the convolutions in TCN residual blocks with transpose convolutions, for the decoder. \textbf{LSTM Variational Auto-Encoder (LSTM VAE)} models the data generating process from the latent space to the observed space and is trained using variational techniques \cite{Doersch2016TutorialOV, park2018lstmvae}. Lastly, the recently proposed \textbf{MSCRED} \cite{zhang2019deep} learns to reconstruct signature matrices, i.e. matrices representing cross-correlation relationships between channels constructed by pairwise inner-product of the channels. Due to the high dimensionality of WADI, we compress WADI to 90\% explained variance by PCA before applying MSCRED. 

In addition to AEs, we include other generic feature normality learning techniques.  \textbf{BeatGAN} \cite{zhou2019beatgan} uses a Generative Adversarial Network (GAN) framework where reconstructions produced by the generator are regularized by the discriminator instead of fixed reconstruction loss functions. \textbf{NASA LSTM} is a 2-layer LSTM model that uses predictability modeling, i.e., forecasting for anomaly detection \cite{hundman2018detecting}. \cite{hundman2018detecting} also proposed a scoring function and threshold for this model which we test separately as \textbf{NASA LSTM Non-Parametric Thresholding (NPT)}, consisting of the exponentially weighted moving average (EWMA) of the root-mean-square sensor prediction errors as the scoring function, and a non-parametric threshold (NPT) with pruning. 
\vspace{-1mm}
\subsection{Models with predefined scoring functions}
\label{sec: predefined-scoring}
\subsubsection{Anomaly measure-dependent feature learning models}
We test \textbf{Deep Auto-Encoding Gaussian Mixture Model (DAGMM)}\cite{zong2018deep}, an algorithm for unsupervised multivariate anomaly detection (that can also be used in the semi-supervised setting) where an AE is trained end-to-end with the scoring function - a Gaussian Mixture Model (GMM), fitted over the concatenation of the hidden space and summary metrics from the reconstruction, that returns the anomaly score $\mathbf{a}$ directly. We also test \textbf{OC-SVM}\cite{manevitz2002ocsvm}, a classic shallow technique for unsupervised anomaly detection via one-class classification. OC-SVM learns the hyperplane encompassing normal samples, and returns a binary classification label based on the side of the separating hyperplane that the sample is on. To reduce computational complexity of OC-SVM, we compress the samples by PCA retaining 90\% explained variance. 
\subsubsection{End-to-end anomaly scoring algorithms} 
\textbf{One-Class Adversarial Nets, OCAN} \cite{zheng2019ocan} is an end-to-end one-class classification method. Here a generator is trained to produce examples \textit{complimentary} to healthy patterns, which is used to train a discriminator for anomaly detection via the GAN framework. \textbf{OmniAnomaly}\cite{su2019omni} is a prior-driven stochastic model for MVTS anomaly detection that directly returns the posterior reconstruction probability of the MVTS input. The log of the probability serves as the channel-wise score, which is summed across channels to get the anomaly score. 
\vspace{-1mm}
\subsection{Scoring functions}
\label{sec: scoring}
We use 4 scoring functions for models in section \ref{sec: independent-scoring}. 
\textbf{Normalized errors}: Errors, $\mathbf{Er}_t^i$ can be used directly as the anomaly score \cite{munir2018deepant}. In order to account for differences in the training error across channels, we subtract the channel-wise mean training reconstruction error from the test errors, before taking the root-mean-square across channels.

\textbf{Gauss-S}:
{Similar to \cite{malhotra2016lstm}, we fit a Gaussian distribution to the training errors and design a score based on the fitted distribution. We note that directly using the probability distribution function (pdf) as in $-\log pdf$ would give points at both tails of the distribution high scores. In particular, this means that even points with very low reconstruction errors will be classified as being anomalous. To avoid this, we instead develop a score based on the cumulative distribution function (cdf): $-\log (1 - cdf)$, that increases monotonically with reconstruction error as $f_{trans}$. To obtain the final anomaly score $\mathbf{a}_t$, we simply add the channel-wise scores \cite{su2019omni}, assuming independence, as suggested by\cite{ahmad2017numenta}. Formally, with
$\hat{\mu}^i, \hat{\sigma}^i$ the empirical mean and standard deviation respectively of the channel-wise errors, and $\Phi$ the cdf of $N(0, 1)$,
\vspace{-1mm}
\begin{equation}
\label{eqn:gauss-s}
\mathbf{A}_t^i = -\log{(1 - \Phi{(\frac{\mathbf{Er}_{t}^i-\hat{\mu}^{i}}{\hat{\sigma}^i})})} \quad\text{;}\quad 
\mathbf{a}_{t} = \sum_{i=1}^{m} {\mathbf{A}^{i}_{t}}
\end{equation}

\textbf{Gauss-D}: In the dynamic gaussian scoring function, we replace the static mean and variance of Gauss-S in equation \ref{eqn:gauss-s} with dynamic mean and variance, $\hat{\mu}^i_t$ and $\hat{\sigma}^i_t$, in order to adapt better to long-term changes occurring during the testing phase \cite{ahmad2017numenta}. 
\begin{equation}
    \label{eqn:dyn-mean}
    \hat{\mu}^i_t = \frac{1}{W}\sum_{j=0}^{W-1}{\mathbf{Er}_{t-j}^i} \hspace{0.5em}\text{;}\hspace{0.5em}
    (\hat{\sigma}^i_t)^2 = \frac{1}{W-1}\sum_{j=0}^{W-1}{({\mathbf{Er}_{t-j}^i - \hat{\mu}^i_t})^2}
\end{equation}
\noindent
where $W$ is the window size. We prepend the last $W-1$ values from the training set to the test set in order to compute $\hat{\mu}^i_t$ and $\hat{\sigma}^i_t$ for $t<W$, the initial part of test. 

\textbf{Gauss-D-K}: This scoring function refers to Gaussian kernel convolution applied on top of Gauss-D\cite{ahmad2017numenta}. The smoothing of the score before aggregation can potentially amplify the total anomaly score even when multiple channels respond to an anomaly at slightly different times, unlike the previous three scoring functions. $\mathbf{A}^i_t$ for the Gauss-D-K scoring function is: 

\begin{equation}
    G(u;\sigma_k) = e^{- \frac{1}{2}(\frac{u}{\sigma_k})^2} \quad \text{;} \quad
    \mathbf{A}_t^i = G * \mathbf{A}^i_{t, Gauss-D}
\end{equation}
\noindent
where $G$ is a Gaussian filter with kernel sigma $\sigma_k$, $*$ the convolution operator and $\mathbf{A}^i_{t, Gauss-D}$ the channel-wise anomaly scores from Gauss-D. $\mathbf{a}_{t}$ is the same as that in equation \ref{eqn:gauss-s}. 

\subsection{Thresholding functions}
\label{sec:thresholding}
A validation set with anomalies could be used to set a static threshold \cite{malhotra2016lstm, zhang2019deep} but we do not assume the availability of such a validation set in our study. We use the following three thresholding functions to evaluate algorithms: 

\textbf{Best-F-score}: This is the static threshold that results in the maximum value of the desired metric ($F_1$, point-adjusted $F_1$ or $Fc_1$ ) for a given $\mathbf{a}_{t}$. We use the best-F-score threshold to get an upper limit on the anomaly detection performance of an algorithm for a static threshold, similar to \cite{su2019omni, li2019mad,audibert2020usad}, and also for comparison of various metrics in section \ref{sec:metrics_compare}. However this requires use of labels in the test set, so it is not applicable in practice. 

\textbf{Top-k}\cite{zong2018deep}: The top-k threshold is the threshold that results in exactly k time-points being labeled as anomalous, where k is the actual number of anomalies in the test set\footnote{We set a separate top-k threshold for each entity.}. We use this threshold to compare across algorithms since k is constant across algorithms. However since the top-k threshold requires anomaly scores for the full test set before thresholding, it is only applicable to a non-streaming scenario. 

\textbf{Tail-p}: When the anomaly scores correspond to probabilities, the tail-p threshold labels scores $\mathbf{a}_{t}<\epsilon$ as anomalous, where $\epsilon$ is a small tail probability \cite{ahmad2017numenta}. This is applicable to Gauss-S, Gauss-D and Gauss-D-K scoring functions in a streaming scenario. Since our scoring function for MVTS is a sum of $m$ negative log probabilities, we set the threshold as $th_{tail-p} = -m\log_{10}(\epsilon)$ (except SMAP and MSL, where $th_{tail-p} = -\log_{10}(\epsilon)$ based only on the single sensor channel error). In practice, $\epsilon$ may be tuned during the initial part of testing, therefore, here we test 5 different values of $-\log_{10}(\epsilon) \in \{1, 2, 3, 4, 5\}$ for each algorithm and dataset pair, and report the highest score obtained. For the multi-entity datasets, we pick a single best value of $-\log_{10}(\epsilon)$ to be applied to all entities. 

\section{Evaluation Metrics}
\begin{figure}[t]
    \centering
    \includegraphics[width=\linewidth]{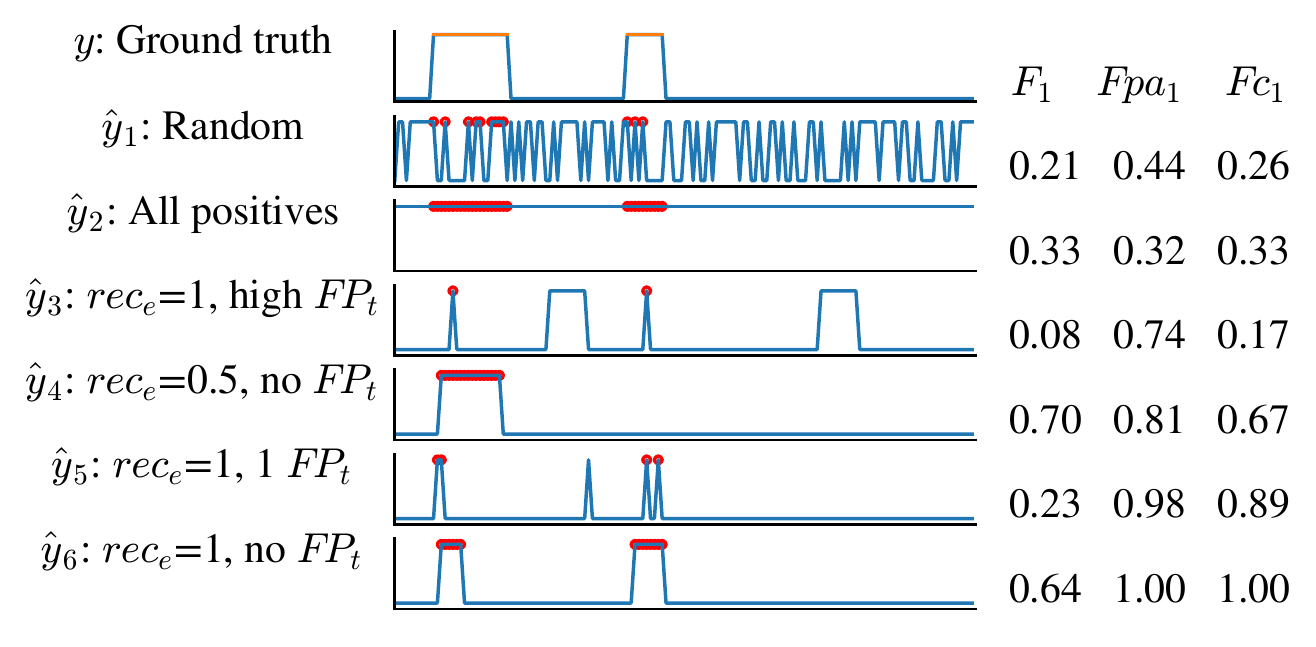}
    \caption{Synthetic ground-truth anomaly labels (topmost, y), with anomalies denoted as 1 (orange), followed by 6 sets of predicted labels, $\hat{y}_1$ through $\hat{y}_6$. The y-axis range is 0-1, and circles (in red) indicate TP time-points on each plot. The scores from each metric are on the right side.}
    \label{fig:example_scores}
\end{figure}

\begin{table*}[t]
\centering
\caption{Metrics with the best-F-score threshold, comparing Random Anomaly Detector (RAD) against UAE (with Gauss-D scoring). Bold values indicate cases where a metric ranks RAD higher than UAE.}
\label{tab:rand_pa}
\resizebox{\textwidth}{!}{%
\begin{tabular}{l|rr|rr|rr|rr|rr|rr|rr}
\toprule
Metric & \multicolumn{2}{c|}{DMDS} & \multicolumn{2}{c|}{MSL} & \multicolumn{2}{c|}{SKAB} & \multicolumn{2}{c|}{SMAP} & \multicolumn{2}{c|}{SMD} & \multicolumn{2}{c|}{SWaT} & \multicolumn{2}{c}{WADI} \\
{} & RAD & UAE & RAD & UAE& RAD & UAE & RAD & UAE & RAD & UAE & RAD & UAE & RAD & UAE \\
\midrule
Point-wise $F_1$&  0.0293 &  0.5311 & 0.2115 & 0.4514 & 0.5369 & 0.5375 &  0.2055 & 0.3898 & 0.0819 & 0.4351 & 0.0888 & 0.4534 &  0.1090 & 0.3537 \\
Point-adjusted $F_1$ &  0.6371 &  0.7234 & 0.8512 & 0.9204 & \textbf{0.9858} & 0.9695 & 0.7418 &  0.8961 & 0.7585 &  0.9723 & \textbf{0.9170} &  0.8685 & \textbf{0.9613} & 0.9574 \\
$Fc_1$    &  0.0345 & 0.6719 & 0.3242 &  0.7132 & 0.5444 & 0.5612 & 0.2895 & 0.8088 & 0.1067 & 0.8325 & 0.1215 & 0.6953 & 0.1317 & 0.5303 \\
\bottomrule
\end{tabular}
}
\end{table*}
\subsection{Composite F-score}
\label{sec:fc_score}
Anomaly detection in practice needs a threshold to get $\hat{y}$. Some authors \cite{park2018lstmvae, kieu2019ensembleaed} evaluate anomaly detection for all possible thresholds using area under the precision-recall curve (AU-PRC) or the receiver operating characteristic (AU-ROC), but it is more important in practical applications to have a high F-score for a chosen threshold \cite{xu2018donut}. The point-wise F-score ($F_1$) \cite{zhang2019deep,li2019mad, audibert2020usad, park2018lstmvae} is the simplest, but in practice, operators care more about detecting events, i.e. a continuous set of anomalous time-points, rather than individual time-points \cite{xu2018donut}. The event-wise F-score, proposed by \cite{hundman2018detecting} takes events into account, but since it counts only 1 false-positive (FP) for a continuous set of time-points, it rewards a detector that labels the entire test-set as a single anomalous event. The point-adjusted F-score ($Fpa_1$) proposed by\cite{xu2018donut}, and used by \cite{su2019omni, audibert2020usad} also accounts for events, but it might give a high score even when a large number of events are not detected\cite{audibert2020usad}. In the $Fpa_1$ score, if any time-point within an event is a true positive (TP), all the time-points within that event are counted as TPs, and then a point-wise F-score is calculated. 

An ideal detector would be one that detects at least one time-point in each event and has no FPs. Early detection is also desirable \cite{ahmad2017numenta}, but is not the subject of our study and has not been considered in any of the works that we compare against. Thus, a good metric for MVTS anomaly detection should 1. reward high precision and high event-wise recall, 2. give a useful indication of how far we are from the ideal case. Following directly from the first criterion, we propose a new metric for time-series anomaly detection - the Composite $F_1$ score, $Fc_1$. $Fc_1$ is the harmonic mean of the time-wise precision, $Pr_t$ and event-wise recall, $Rec_e$: 
\begin{equation}
    Pr_t = \frac{TP_t}{TP_t + FP_t}
\quad\text{and}\quad
    Rec_e = \frac{TP_e}{TP_e + FN_e}
\end{equation}
where $TP_t$ and $FP_t$ are the number of TP and false positive (FP) time-points respectively, while $TP_e$ and $FN_e$ are the number of TP and false negative (FN) events respectively. $TP_e$ is the number of true events for which there is at least one TP time-point. Remaining true events are counted under $FN_e$. 
\subsection{Comparison of metrics for anomaly detection}
\label{sec:metrics_compare}
To gain intuition on what each metric rewards in a labeling scheme, we compare the three F-scores - $F_1$, $Fpa_1$ and $Fc_1$ for predicted labels on a synthetic test case shown in Fig. \ref{fig:example_scores}. All three metrics give low scores to the spurious labeling schemes $\hat{y}_1$ and $\hat{y}_2$, and all metrics suggest that $\hat{y}_4$ is better than $\hat{y}_1$-$\hat{y}_3$. However, the $Fpa_1$ score for $\hat{y}_3$ is 0.74, which suggests good performance in spite of high FPs. $\hat{y}_5$ is highly desirable due to high $rec_e$ and low FPs but the $F_1$ score is low. $\hat{y}_6$ is an ideal detector but $F_1$ ranks it lower than $\hat{y}_4$. Each metric ranks these labeling schemes differently, but $Fc_1$ is the only one that gives low scores (e.g. $<0.6$) to the undesirable predictions $\hat{y}_1$-$\hat{y}_3$ and higher scores to the desirable predictions $\hat{y}_4$-$\hat{y}_6$, while correctly suggesting that $\hat{y}_6$ is ideal.

Furthermore, in order to establish baseline values of various F-score metrics on each dataset, we evaluate the performance of a trivial detector - the Random Anomaly Detector (RAD) with the best-F-score threshold (Table \ref{tab:rand_pa}). RAD simply assigns a random real score in $[0,1]$ for each time-point. The $Fpa_1$ score suggests good performance ($Fpa_1>0.6$) for RAD on all datasets, while the $F_1$ and $Fc_1$ behave as expected, showing low scores. Table \ref{tab:rand_pa} also shows results for UAE with Gauss-D scoring, which is the best overall algorithm for anomaly detection in our study. Surprisingly, the $Fpa_1$ score suggests that RAD is \textit{better} than UAE on SKAB, SWaT and WADI. RAD's $Fpa_1$ score of 0.9613 on WADI is attained at a threshold that labels only 0.17\% time-points as anomalous, so that the number of FPs is kept low, but the number of TPs is exaggerated as the noisy scoring function can label at least 1 time-point in most events. Thus, Figure \ref{fig:example_scores} and Table \ref{tab:rand_pa} show that the $F_1$ score may be too pessimistic, and the $Fpa_1$ may be too optimistic in some cases. 

\subsection{Metrics for anomaly diagnosis}
For anomaly diagnosis, we assume that the ground-truth of anomalous time-points is known but the causes are unknown. We use HitRate@150 \cite{su2019omni} and the Root Cause-top-3 (RC-top-3) metrics to evaluate anomaly diagnosis. The HitRate@150\cite{su2019omni} metric gives the average fraction of overlap between the true causes and the top $1.5c$ identified causes, where $c$ is the number of true causes. The RC-top-3 (or RC-top-k) is a new metric we propose, which gives the fraction of events for which at least one of the true causes was identified to be in the top 3 (or top k) causes identified by the algorithm. HitRate@150 rewards identifying \textit{all} of the true causes while RC-top-3 rewards identifying at least one of the causes.
\section{Experimental setup}
\label{sec:experiments}
\textbf{Dataset Pre-processing}: We carry out channel-wise min-max normalization for each channel squashing it in the range of [0, 1] during train and clipping it to [min-4, max+4], i.e.the range [-4, 5] during test to prevent excessively large values from a particular channel skewing the overall scores. The window length ($l_w$) and step size ($l_s$) are shown in Table \ref{tab:datasets}. We choose $l_w=100$, similar to \cite{su2019omni} in the absence of specific knowledge of the time-scales in the datasets, but choose a smaller $l_w=30$ for WADI due to its higher dimensionality. We choose $l_s = 1$ for SKAB, MSL, SMAP and SMD datasets as they are shorter and $l_s = 10$ for the remaining, longer datasets to speed up training. 

\textbf{Hyperparameters and Implementation}: 
When available, we adapt pre-existing implementations (Table \ref{tab:code-refs} in SI). We tune the hyperparameters of LSTM-VAE, TCN AE and FC AE models using random search over the search space shown in the SI, Table \ref{tab:hyperparam_tuning} choosing the configuration with the minimum reconstruction error on the validation set, as done by \cite{malhotra2016lstm}. We set the hyperparameters empirically for the remaining algorithms, summarized in SI Table \ref{table:hyperparam}. An extensive hyperparameter tuning for each model and dataset is beyond the scope of this work. Runtimes of the models on 3 datasets are provided for reference in SI Table \ref{tab:train-times}. We choose $W$ for Gauss-D and Gauss-D-K scoring functions to be comparable to training set size, W=100000 for SWaT, WADI and DAMADICS, W=2000 for MSL and SMAP, W=25000 for SMD and W=100 for SKAB. We set $\sigma_k$ for Gauss-D-K scoring empirically to 120 for SWaT and WADI, 5 for DAMADICS, 10 for SMAP and MSL, and 1 for SMD and SKAB. 

\textbf{Evaluation}: We use $Fc_1$ score with the top-k threshold as the primary metric in our anomaly detection evaluation. For completeness, we report additional results for $Fc_1$ score (SI sections \ref{sec:supp-fc-best} and \ref{sec:supp-fc1-unsup}) with best-F-score and tail-p thresholds respectively, the $F_1$ score with the best-F-score threshold (SI section \ref{sec:supp-f1}), AU-ROC score (SI section \ref{sec:supp-auroc}), and AU-PRC score (SI section \ref{sec:supp-auprc}). For anomaly diagnosis we report the RC-top-3 scores (main manuscript) and HitRate@150 (SI section \ref{sec:supp-rc}). 
For multi-entity datasets, anomaly detection and diagnosis metrics are averaged over entities. We run each experiment 5 times with different seeds and report the average. We use the recommendations by \cite{demvsar2006statistical} to compare statistical significance at significance level $\alpha=0.05$. We conduct overall comparison using the Friedman test\cite{friedman1937use}, under the null hypothesis that all methods perform the same. If we can reject the null hypothesis, we conduct post-hoc tests using Hochberg's step-up procedure\cite{hochberg1988sharper}, comparing the best-performing method against all others. The details of the tests are described in SI section \ref{sec:statistical-tets}. 
\vspace{-1mm}
\section{Results}
\label{sec:results}
\subsection{Anomaly Detection}
\label{sec: ano-det-results}
Fig. \ref{fig:scoring_funcs_topk} shows dataset-wise effect of the choice of model (colors and symbols), and scoring function (X-axis) on the $Fc_1$ performance for the top-k threshold. We see that both the model and scoring function can make a significant difference to the $Fc_1$ score. 
\begin{figure}[h]
\centering
    \includegraphics[width=1.0\linewidth]{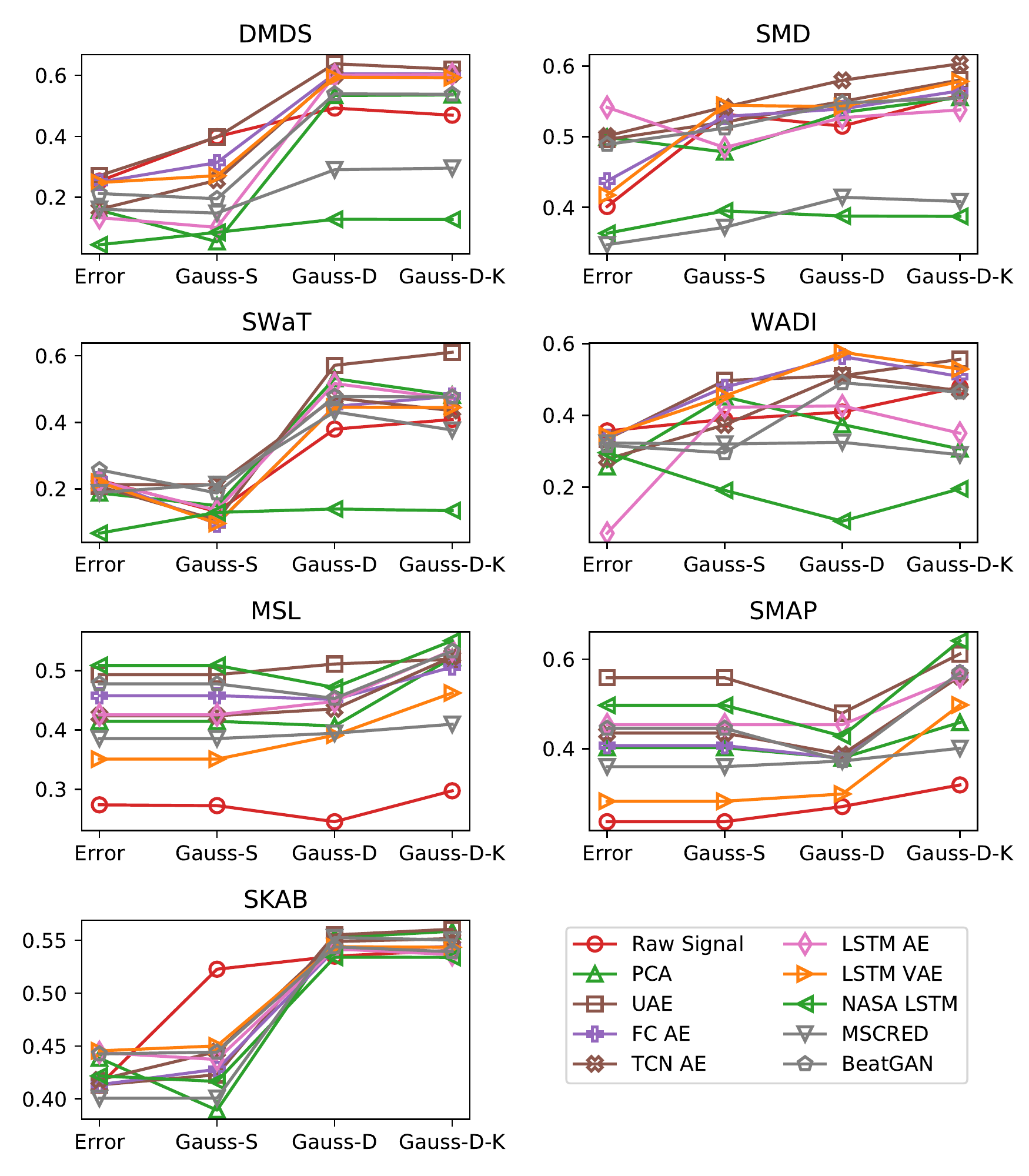}
    \caption{Effect of scoring functions on the $Fc_1$ performance with top-k threshold for various models and datasets. Results in this plot are also provided as a supplemental csv file.}
    \label{fig:scoring_funcs_topk}
\end{figure}
\begin{table}[h]
\setlength{\tabcolsep}{4pt}
\caption{Average $Fc_1$ score, with the Gauss-D scoring function (except those denoted with *) and top-k threshold. See SI Table \ref{tab: fc-top-k-std} for standard deviation.}
\label{tab: fscore-top-k}
\resizebox{\linewidth}{!}{%
\begin{tabular}{p{0.18\linewidth}rrrrrrr|p{0.08\linewidth}p{0.05\linewidth}}
\toprule
{} &    DMDS &     MSL &    SKAB &    SMAP &     SMD &    SWaT &    WADI &    Mean &  Rank \\
\midrule
Raw Signal  &  0.4927 &  0.2453 &  0.5349 &  0.2707 &  0.5151 &  0.3796 &  0.4094 &  0.4068 &       9.3 \\
PCA         &  0.5339 &  0.4067 &  0.5524 &  0.3793 &  0.5344 &  0.5314 &  0.3747 &  0.4733 &       5.6 \\
UAE         &  \textbf{0.6378} &  \textbf{0.5111} &  \textbf{0.5550} &  \textbf{0.4793} &  0.5501 &  \textbf{0.5713} &  0.5105 &  \textbf{0.5450} &       \textbf{1.6} \\
FC AE       &  0.6047 &  0.4514 &  0.5408 &  0.3788 &  0.5395 &  0.4478 &  0.5639 &  0.5038 &       4.7 \\
LSTM AE     &  0.5999 &  0.4481 &  0.5418 &  0.4536 &  0.5271 &  0.5163 &  0.4265 &  0.5019 &       4.7 \\
TCN AE      &  0.5989 &  0.4354 &  0.5488 &  0.3873 &  \textbf{0.5800} &  0.4732 &  0.5126 &  0.5052 &       3.9 \\
LSTM VAE    &  0.5939 &  0.3910 &  0.5439 &  0.2988 &  0.5427 &  0.4456 &  \textbf{0.5758} &  0.4845 &       6.0 \\
BeatGAN     &  0.5391 &  0.4531 &  0.5437 &  0.3732 &  0.5479 &  0.4777 &  0.4908 &  0.4894 &       5.0 \\
MSCRED      &  0.2906 &  0.3944 &  0.5526 &  0.3724 &  0.4145 &  0.4315 &  0.3253 &  0.3973 &       8.1 \\
NASA LSTM   &  0.1284 &  0.4715 &  0.5339 &  0.4280 &  0.3879 &  0.1398 &  0.1058 &  0.3136 &       8.9 \\
DAGMM*       &  0.0000 &  0.1360 &  0.0000 &  0.1681 &  0.0187 &  0.0000 &  0.0256 &  0.0498 &      12.9 \\
OmniAnomaly* &  0.1425 &  0.4120 &  0.4561 &  0.3767 &  0.5002 &  0.1466 &  0.2443 &  0.3255 &       9.4 \\
OCAN*        &  0.2532 &  0.3009 &  0.4369 &  0.2787 &  0.4614 &  0.1547 &  0.0000 &  0.2694 &      11.0 \\
\bottomrule
\end{tabular}
}
\end{table}
\subsubsection{Effect of Scoring Function}
For DMDS, SMD, SWaT and SKAB, the choice of scoring function in general makes a bigger difference than the choice of model. For example, the $Fc_1$ score for the UAE model on SWaT ranges from $\sim$0.1-0.6 for different scoring functions, while the performance of all models (except NASA LSTM) for the Gauss-D-K scoring function is in the range $\sim$0.4-0.6. The ranks of the scoring functions, averaged across datasets and models shown in Fig \ref{fig:scoring_funcs_topk} are: Gauss-D-K - 1.5, Gauss-D - 2.0, Gauss-S - 3.2 and Error - 3.3. Statistical tests indicate that the differences between the performance of the dynamic scoring functions - Gauss-D-K, Gauss-D vs. the static ones - Gauss-S, Error are statistically significant (see SI section \ref{sec:stat-detection-scoring}). 

The dynamic scoring functions outperform the static scoring functions as they adapt to the changing normal during test, but they can also adapt to anomalies in the test set. This does not affect the overall score much when the score is aggregated across channels and when the Gauss-D window length, W is large compared to anomalous event lengths. However for SMAP and MSL, no aggregation of scores takes place, and for SMAP the average event length (1001 mins) is comparable to W (2000 mins), which may be why Gauss-D performs worse than Gauss-S on SMAP. 

Gauss-D-K performs better overall than Gauss-D by amplifying anomaly scores from multiple channels responding to the same anomaly, even if they respond at slightly different times. However, Gauss-D-K performance is sensitive to the value of $\sigma_k$ which is not straightforward to set, and we set it empirically here in a non-rigorous search to approximately attain good test performance. In contrast, the window size, W for Gauss-D was set simply as a size comparable to the training set size. For this reason, we choose the Gauss-D scoring function to compare the effect of the choice of model on anomaly detection. 

\subsubsection{Effect of Model}
\begin{figure}[h]
    \includegraphics[width=0.9\linewidth]{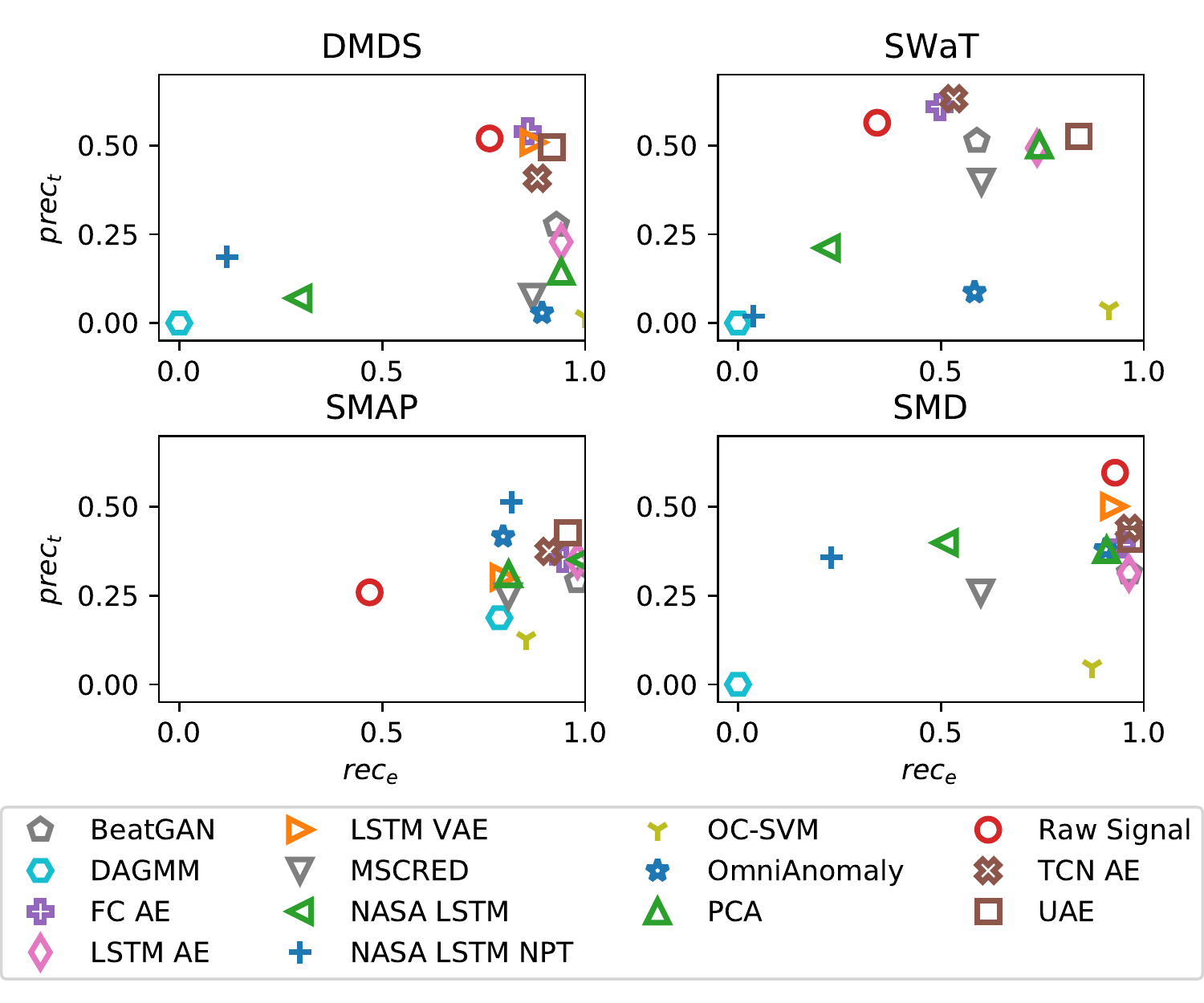}
    \caption{Plots of $prec_t$ vs. $rec_e$ for algorithms with the top-k threshold and scoring functions as in Table \ref{tab: fscore-top-k}. See additional datasets in SI Figure \ref{fig:supp_prec_rec_topk}.}
    \label{fig:prec_rec_topk}
\end{figure}
Table \ref{tab: fscore-top-k} summarizes the $Fc_1$ score for models using either Gauss-D scoring or predefined scoring functions (starred), with the top-k threshold. We see that UAE is the best performing model for five out of seven datasets, and is leading overall by mean $Fc_1$ and average rank. We find that the difference between the performance of UAE against Raw Signal and several recently proposed algorithms - DAGMM\cite{zong2018deep}, OCAN\cite{zheng2019ocan}, OmniAnomaly\cite{su2019omni}, NASA LSTM\cite{hundman2018detecting} and MSCRED\cite{zhang2019deep} is statistically significant, while the remaining comparisons are statistically insignificant (see SI section \ref{sec:stat-detection-model}), perhaps due to the large number of algorithms tested with just 7 datasets. 
UAE is also the top performing in  Tables \ref{tab: top-k-gauss-d-k}, \ref{tab: f1_bestf1_gauss-d}, \ref{tab: f1_bestf1_gauss-dk},  \ref{tab: auroc-gauss-d}, \ref{tab: auroc-gauss-d-k}, \ref{tab: auprc-gauss-d}, and \ref{tab: auprc-gauss-d-k} in the SI where we compare models under other scoring functions (Gauss-D-K) or other metrics - point-wise $F_1$, AU-ROC and AU-PRC. 

The superior performance of UAE could be attributed to its channel-wise models that learn and retain information about each channel before aggregating scores across channels, enabling it to effectively detect temporal anomalies which are prevalent in these datasets. Another question is why do other sophisticated techniques that model both temporal and multivariate correlations perform worse than UAE? The reason may be that in a dataset which has large temporal and inter-channel correlations (typical of the datasets we use), the two effects can get confounded, resulting in spurious correlations (e.g. SI Fig \ref{fig:omni_spurious}). The space of possible anomalies is also much larger when both cross-channel and temporal anomalies are considered by the algorithm, which might result in more noisy anomaly detection behavior. 

The second best performing model by average rank is FC AE. Interestingly, FC AE and TCN AE perform better than RNN based methods - LSTM AE, LSTM VAE and OmniAnomaly, even though RNN based methods are deemed to be better suited to time-series tasks. This may be because the fixed length time-series can easily be modeled by FC and TCN units as well. The poor performance of OmniAnomaly\cite{su2019omni}, DAGMM\cite{zong2018deep} and OCAN\cite{zheng2019ocan} could be attributed partly to the use of static scoring functions. The forecasting model NASA LSTM performs poorly for all scoring functions in Fig. \ref{fig:scoring_funcs_topk} for multiple datasets, suggesting that auto-encoder based techniques might be better than forecasting techniques for semi-supervised MVTS anomaly detection. 

Fig. \ref{fig:prec_rec_topk} shows the $prec_t$ and $rec_e$ values for the algorithms presented in Table \ref{tab: fscore-top-k} with the top-k threshold. While the best algorithms are able to attain a high event-wise recall, time-wise precision for every dataset and algorithm is below 0.5. This suggests that there might exist better thresholds that tag fewer anomalies, which could improve $prec_t$ without lowering $rec_e$ too much; Table \ref{tab: fc_bestf1} in SI shows $Fc_1$ score with the Gauss-D scoring function and the best-F-score (best-$Fc_1$ in this case) threshold, and these scores are generally higher than the top-k threshold scores shown in Table \ref{tab: fscore-top-k}.

Table \ref{tab: fscore unsup} in the SI shows $Fc_1$ scores for all datasets and algorithms with the Gauss-D scoring function and tail-p threshold, or a different streaming threshold as noted in Table \ref{tab: fscore unsup}. The tail-p threshold results show similar trends to the top-k threshold results of Table \ref{tab: fscore-top-k}, with UAE leading the pack based on overall mean and average rank. The precision-recall results in the SI Fig. \ref{fig:prec_rec_unsup} show that with the streaming threshold, different techniques pick different trade-offs between $prec_t$ and $rec_e$, and higher values of $prec_t$ are attained than those shown in Fig. \ref{fig:prec_rec_topk}. 
\subsection{Anomaly Diagnosis Results}
\begin{figure}[h]
\centering
    \includegraphics[width=0.9\linewidth]{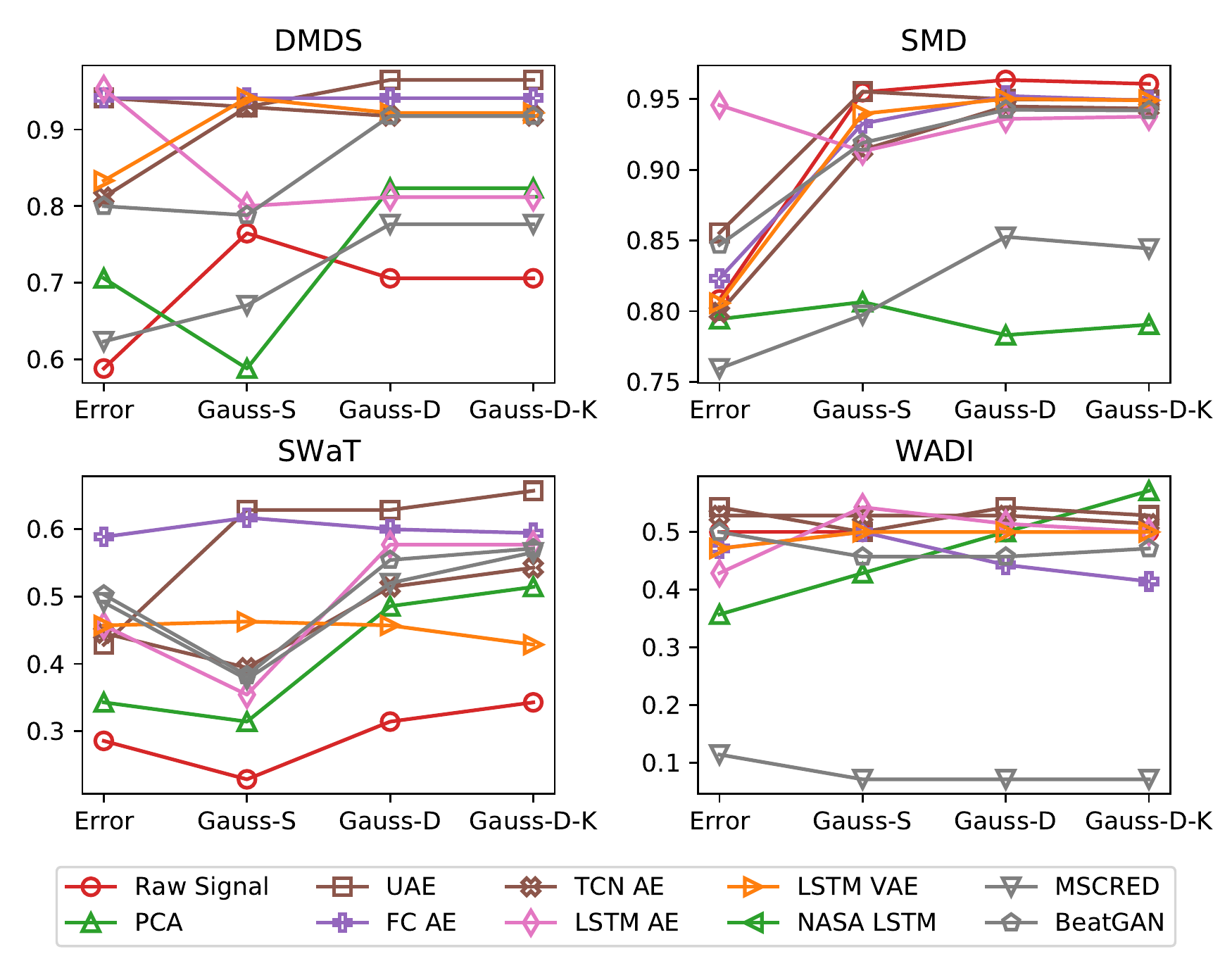}
    \caption{Effect of scoring functions on the RC-top-3 performance for various algorithms and datasets.}
    \label{fig:scoring_funcs_rctop3}
\end{figure}
\begin{table}[h]
\centering
\setlength{\tabcolsep}{4pt}
\caption{RC-top-3 metric using the Gauss-D scoring function (except starred). See SI Table \ref{tab: rc_top3_all_std} for standard deviation.}
\label{tab: rc_top3_all}
\resizebox{\linewidth}{!}{%
\begin{tabular}{lrrrrp{0.1\linewidth}p{0.1\linewidth}}
\toprule
Algo & DMDS & SMD & SWaT & WADI & Overall mean & Avg Rank\\
\midrule
Raw Signal            &   0.7059 &  \textbf{0.9635} &  0.3143 &    0.5000 &      0.6209 & 6.5\\
PCA                   &   0.8235 &  0.7831 &  0.4857 &   0.5000 &         0.6481 & 7.2\\
UAE &   \textbf{0.9647}  &  0.9498 &  \textbf{0.6286}  &  \textbf{0.5428}  &       \textbf{0.7715} & \textbf{1.8}\\
FC AE        &   0.9412 &    0.9522  &  0.6000  &  0.4428  &       0.7341 & 3.8 \\
LSTM AE      &   0.8117  &  0.9360  &  0.5771  &  0.5143  &       0.7098 & 5.2 \\
TCN AE       &   0.9177 &    0.9448 &  0.5143  &  0.5286  &       0.7263 & 4.5 \\
LSTM VAE              &   0.9177 &  0.9501  &  0.4571  &  0.5000 &       0.7062 & 4.8\\
BeatGAN     &  0.9176 &  0.9424 &  0.5543 &  0.4571 &  0.7178 &       5.8 \\
MSCRED      &  0.7765 &  0.8526 &  0.5200 &  0.0714 &  0.5551 &       8.2 \\
OmniAnomaly*           &   0.9177 &   0.9272  &  0.3772 &   0.4857  &       0.6770 & 6.25 \\
\bottomrule
\end{tabular}
}
\end{table}
Fig. \ref{fig:scoring_funcs_rctop3} shows the effect of scoring functions and models on the independent anomaly diagnosis performance using the RC-top-3 metric. We see that for DMDS and SMD, scores of $\sim$0.95 are achieved by the best algorithms, suggesting that ranking channels by anomaly score is an effective strategy for independent anomaly diagnosis. The RC-top-3 scores for SWaT and WADI are lower because in these datasets, for some events, channels other than the original causes can also be affected by an anomalous event, and can get diagnosed as the causes (e.g. SI Fig. \ref{Fig:RC}). The HitRate@150 scores in SI Fig. \ref{fig:scoring_funcs_hr150} are lower than RC-top-3 results since HitRate@150 evaluates the ability to identify all of the root causes, rather than at least one. However, the trends across datasets, models and scoring functions are similar to that seen in Fig. \ref{fig:scoring_funcs_rctop3}. Among scoring functions, Gauss-D has the highest average rank (2.0) but we find the differences between scoring functions to be statistically insignificant.

\textbf{Effect of Model}: Table \ref{tab: rc_top3_all} shows the RC-top-3 results for various models with the Gauss-D scoring function (except OmniAnomaly). Again, UAE is the top performer with an average rank of 1.5.  Similar results are seen in SI Table \ref{tab:hr_150_all} showing the HitRate@150 performance. Thus, even though we evaluate anomaly detection and diagnosis independently, the results from both evaluations are consistent and confirm the intuition that an algorithm that is good at anomaly detection, i.e. ranking the anomaly score correctly across time-points, is also good at anomaly diagnosis, i.e. ranking the scores correctly across channels, before aggregation. However, the overall comparison across models on anomaly diagnosis is not statistically significant, likely due to the large number of comparisons on just 4 datasets. 

\vspace{-1mm}
\section{Conclusions and Future Work}
\label{sec:conclusions}
We conducted a comprehensive evaluation of deep-learning based algorithms on MVTS anomaly detection and independent anomaly diagnosis, by training 11 deep learning models on 7 MVTS datasets (114 entities) and 5 repeats, resulting in 6270 deep learning models and 4273 end to end experiments. We showed through experiments that existing evaluation metrics in use for event-wise time-series anomaly detection are not adequate and can be misleading. To remedy this, we proposed a new metric for event-wise time-series anomaly detection, the composite F-score, $Fc_1$, which is the harmonic mean of event-wise recall and point-wise precision. Unlike previous studies, we studied the effect of models and scoring functions independently to gain a deeper understanding of what makes a good time-series anomaly detection algorithm. We found that the choice of the scoring function can have a large impact on anomaly detection performance, and dynamic scoring functions Gauss-D and Gauss-D-K work better than the static scoring function, Gauss-S. Surprisingly and significantly, we found that the top performing model in our evaluation for anomaly detection and diagnosis was the UAE model with the Gauss-D scoring function. 

While the good performance of UAE could be partly due to the prevalence of temporal anomalies, our study makes it clear that recently proposed deep algorithms \cite{zhang2019deep,zhou2019beatgan,hundman2018detecting,su2019omni,zheng2019ocan} fail to effectively detect even these simple \cite{wu2020current} anomalies in MVTS datasets. The UAE model would be a good starting point for anomaly detection in a steady state CPS dataset, but might not perform as well on a system with multiple operating states. Possible routes to further improve upon the performance of UAE may be through the use of improved channel-wise models and through ensembling the channel-wise reconstructions with reconstructions from a complementary model that accounts for only cross-channel effects, similar to the idea used by \cite{zhang2019gpmultivar}. Our study also highlights the need for more challenging CPS datasets that display multiple operating conditions and where both temporal and cross-channel anomalies are observed. 

Overall, our work provides important insights into the design and evaluation of methods for anomaly detection and diagnosis, and serves as a useful guide for future method development.

\ifCLASSOPTIONcompsoc
  \section*{Acknowledgments}
\else
  \section*{Acknowledgment}
\fi

We would like to thank the HBMS IAF-PP grant H19/01/a0/023 and A*STAR AME Programmatic Funds (Grant No. A20H6b0151) for funding part of this study.

\ifCLASSOPTIONcaptionsoff
  \newpage
\fi



\bibliographystyle{IEEEtran}
\bibliography{bibliography}
%



%
\vspace*{-2\baselineskip}
\begin{IEEEbiography}[{\includegraphics[width=1in,height=1.25in,clip,keepaspectratio]{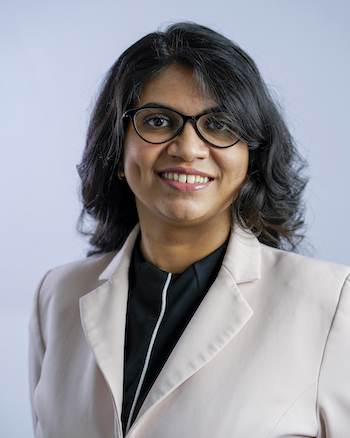}}]{Astha Garg} received her PhD degree in chemical engineering from the Pennsylvania State University, PA, USA in 2017. Her experience spans semiconductor equipment manufacturing (Applied Materials, USA), data-driven chemicals and materials research (Citrine Informatics, USA) and marine equipment monitoring (ChordX Pte. Ltd., current), as well as research on time-series (Institute for Infocomm Research (I2R), A*STAR, Singapore). Her research focuses on applications of machine learning to Industry 4.0, and data-driven experimental design.    
\end{IEEEbiography}
\vspace*{-2\baselineskip}
\begin{IEEEbiography}[{\includegraphics[width=1in,height=1.25in,clip,keepaspectratio]{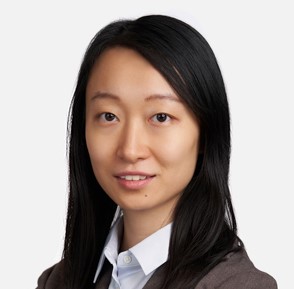}}]{Wenyu Zhang} received the PhD degree in Statistics from Cornell University in 2020. She is currently a Scientist at I2R (Institute for Infocomm Research) working in areas of machine learning and time series analysis.
\end{IEEEbiography}
\vspace*{-2\baselineskip}
\begin{IEEEbiography}[{\includegraphics[width=1in,height=1.25in,clip,keepaspectratio]{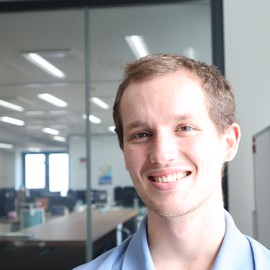}}]{Jules Samaran} is pursuing an integrated undergraduate and Masters degree at Mines Paristech and PSL University, Paris, France as well as Ecole Normale Superieure de Saclay, Paris, France. His research interests lie at the intersection of statistics and machine learning. 
\end{IEEEbiography}
\vspace*{-2\baselineskip}
\begin{IEEEbiography}[{\includegraphics[width=1in,height=1.25in,clip,keepaspectratio]{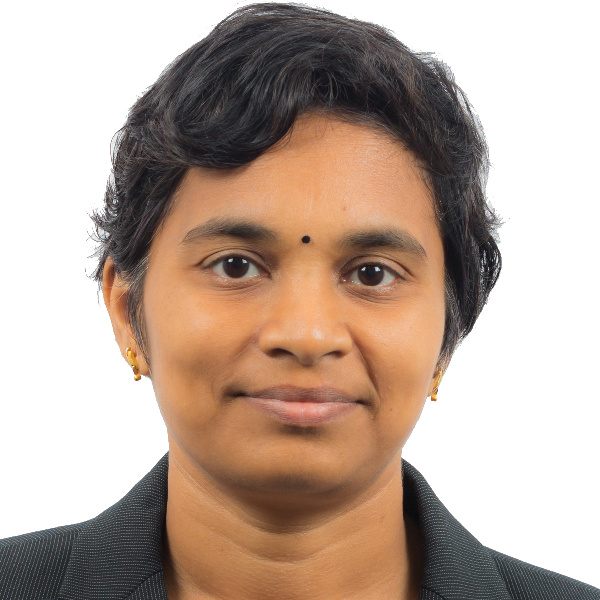}}]{Savitha Ramasamy} received her PhD degree from Nanyang Technological University, Singapore in 2011. Currently, she is a research group leader at Institute for Infocomm Research, A*STAR, Singapore. She has published about 100 papers in various international conferences and journals, along with a research monograph published by Springer-Verlag, Germany. Her research interests are in developing robust AI, with special focus on lifelong learning and time series data analysis, and has led translation of these robust models for predictive analytics in real-world applications. Her contributions to data analysis and AI have been recognized in the inaugural 100 SG Women in Technology list.
\end{IEEEbiography}
\vspace*{-2\baselineskip}
\begin{IEEEbiography}[{\includegraphics[width=1in,height=1.25in,clip,keepaspectratio]{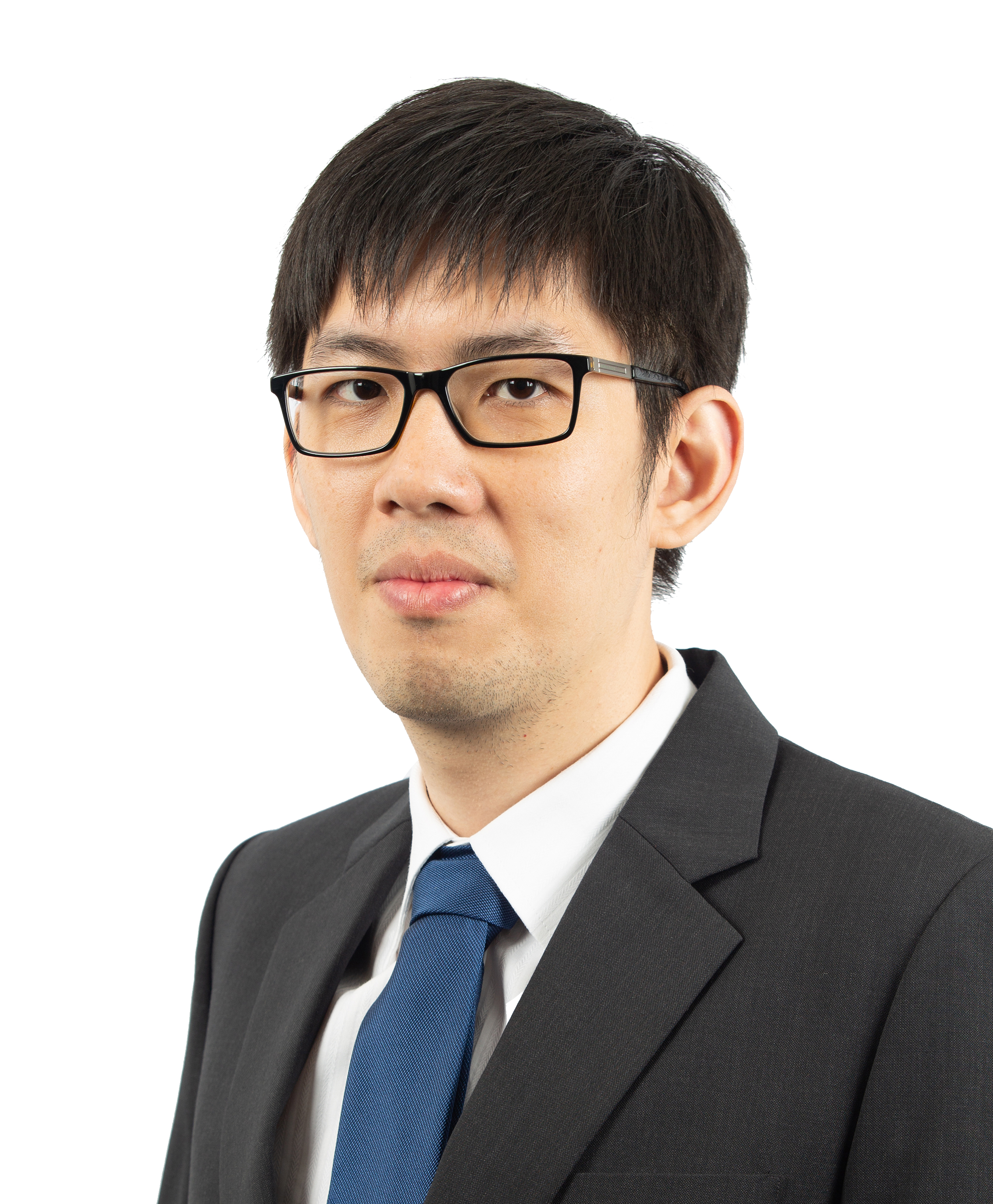}}]{Chuan-Sheng Foo} received his BS, MS and PhD degrees from Stanford University. He currently leads a research group at the Institute for Infocomm Research, A*STAR, Singapore, which focuses on developing data-efficient deep learning algorithms that can learn from less labeled data. 
\end{IEEEbiography}
\clearpage
\newpage

\renewcommand{\labelitemi}{$\bullet$}
\newcommand{\beginsupplement}{
    \setcounter{section}{0}
    \renewcommand{\thesection}{S\arabic{section}}
    \setcounter{equation}{0}
    \renewcommand{\theequation}{S\arabic{equation}}
    \setcounter{table}{0}
    \renewcommand{\thetable}{S\arabic{table}}
    \setcounter{figure}{0}
    \renewcommand{\thefigure}{S\arabic{figure}}
    \newcounter{SIfig}
    \renewcommand{\theSIfig}{S\arabic{SIfig}}}

\onecolumn
\beginsupplement
{\centering \LARGE{Supplemental Information} \par}

\section{Taxonomy of algorithms evaluated}
\label{sec:taxonomy}
\begin{figure}[!h]
    \includegraphics[width=0.8\textwidth]{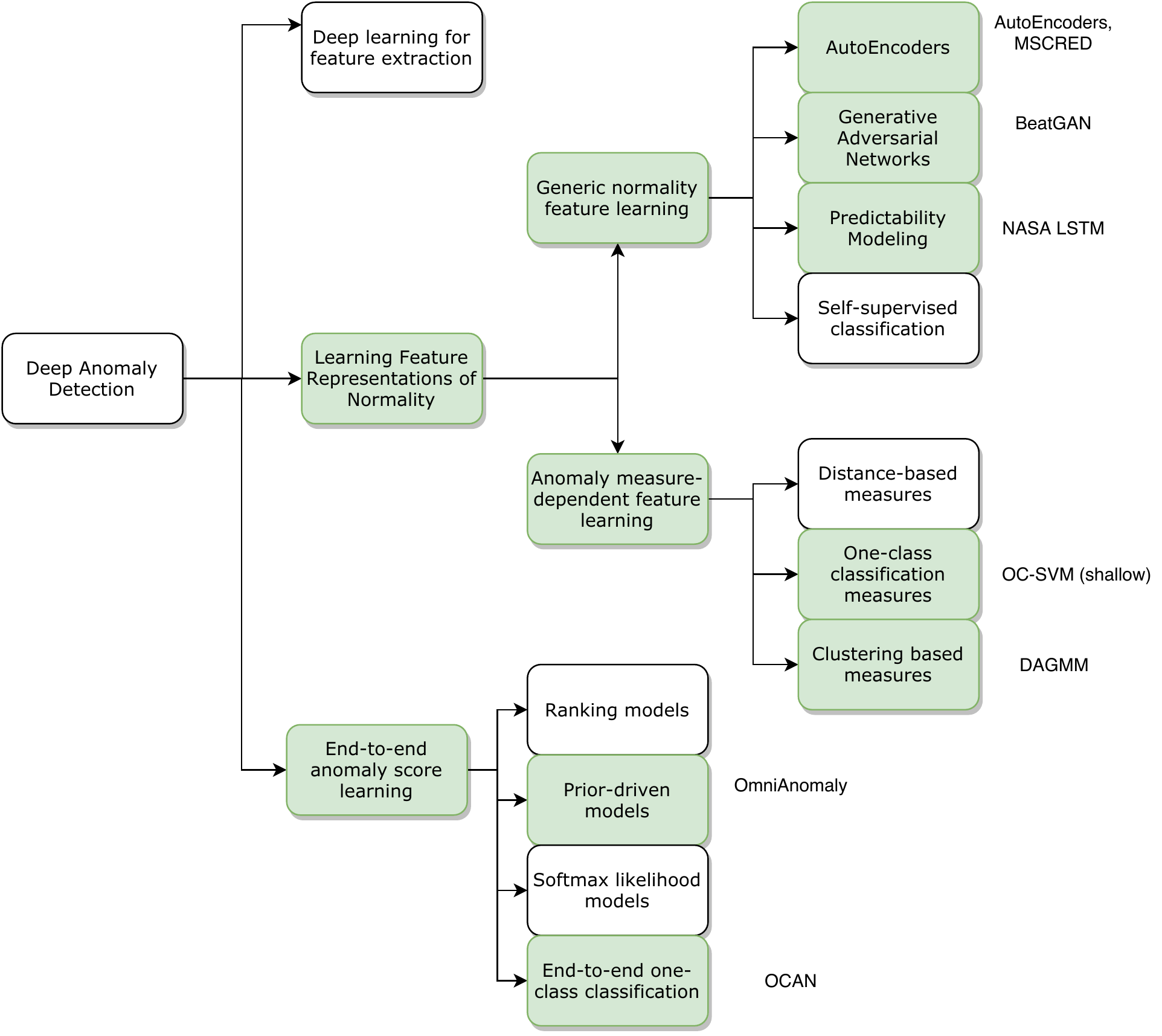}
    \caption{Anomaly detection algorithms tested in this work (shaded green), based on the taxonomy proposed by \cite{pang2021} for deep anomaly detection. While we do include OC-SVM in the schematic, it is not a deep method. In addition to those shown in the diagram, we also test Raw Signal and PCA, which are not deep methods. }
    \label{fig:taxonomy}
\end{figure}

Not all techniques in this taxonomy are suitable for unsupervised or semi-supervised multivariate time-series (MVTS) anomaly detection in a streaming scenario. Based on the discussion on each category in \cite{pang2021}:
\begin{itemize}
    \item \textit{Deep learning for feature extraction} techniques rely on sophisticated feature extraction techniques which were developed primarily for images, and may not be suitable for MVTS.
    \item \textit{Self-supervised classification} techniques rely on data augmentation techniques that have been developed primarily for image data. We are not aware of any existing deep algorithms of this type for MVTS.  
    \item \textit{Distance-based measures} are computationally intensive at test-time, and hence not suitable for streaming test setting that we work in. 
    \item \textit{Ranking models} require some form of labelled anomalies which we do not assume.
    \item \textit{Softmax likelihood models} have been developed primarily for heterogeneuous data sources or categorical data. We are not aware of any existing deep algorithms of this type for MVTS.  
\end{itemize}

\section{Datasets}
\label{sec:supp_ds_info}
The datasets contain data from sensors and actuators which interact with each other and the environment in intelligent and stochastic ways. Due to the dynamic nature of the systems we consider, induced anomalies might only become evident after some delay (sometimes several minutes), and the channel behaviour may not return to normal even after the anomaly inducing mechanism has been withdrawn\cite{goh2016swat}.

\subsection{SWaT}
The Secure Water Treatment dataset \cite{goh2016swat} can be requested on the iTrust website \cite{iTrust}. The dataset has 51 channels, which consist of sensors such as flow meters, level transmitters, conductivity analyzer and actuators such as motorized valves and pumps. The dataset has 14 channels corresponding to signals from various pumps that are constant in train but these are retained as they are allowed to change during testing. The initial 7 days of data consist of normal operation (training set) while 36 attacks were launched in the last 4 days (test set). Each attack compromises one or more channel(s). One of the attacks was much longer (598 mins) than all others (under 30 mins each). This biases the scores of all algorithms depending on whether this event was detected, so we cut this event in the test set to 550 s (the average event length) by discarding anomalous time-points for the rest of the event. Note that since 2 of the attacks were launched right one after the other, we treat it as a single anomalous sequence, and hence consider 35 anomalous events. Of these, root cause labels are available for 33 attacks while the remaining attacks have discrepancies in their root cause labels. 
For one of the attacks, the start and end points did not match any events listed in the `List of Attacks' sheet provided on the dataset website\cite{iTrust}. For another attack, the provided root cause does not match any of the available channels.

\subsection{WADI}
The Water Distribution testbed \cite{ahmed2017wadi} is an operational, scaled-down version of a water distribution network in a city. It is connected to the SWaT plant and takes in a portion of its reverse-osmosis output. The distribution network consists of 3 distinct control processes, namely - primary grid, secondary grid and return water grid - each controlled by its own set of Programmable Logic Controllers (PLCs). The dataset, also hosted by iTrust \cite{iTrust} consists of data from 123 sensors and actuators collected over 14 days for normal operation (training set) and 2 days with 15 attacks (test set). Since 2 attacks were launched at the same time, we consider 14 anomalous events in this paper. We use all 14 events for anomaly detection, but for anomaly diagnosis, we have root cause labels for only 12 of the 14 events. For the remaining 2 events, the labels do not specify exactly which component(s) was compromised. 

\subsection{DMDS}
The DAMADICS (Development and Application of Methods for Actuator Diagnosis in Industrial Control Systems) benchmark \cite{bartys2006damadics} consists of real process data from the Lublin Sugar Factory as well as a simulator to generate artificial faults. Here we use only the real dataset with induced faults in the industrial system, available publicly \cite{damadics}. 
The dataset consists of data for 25 days of operation, from Oct-29 to Nov-22, 2001 of 3 benchmark actuators - one each located upstream and downstream of evaporator station, and the third controlling flow of water to the steam boiler system. Artificial faults were induced on Oct-30, Nov-9, Nov-17 and Nov-20. Unlike SWaT and WADI, the train and test splits for normal and test operation were not provided. We chose train-test splits such that the test is entirely after train as would be expected in a real scenario, the train is continuous, and the train contains no anomalies. Accordingly, we used the data from Nov-3 to Nov-8 (6 days) as the training set and data from Nov-9, Nov-17 and Nov-20 (3 days) as the test set. From this, the first 10800 points from the training set were dropped as the system appeared much more unstable than the rest of the training set, potentially from the anomaly induced earlier on Oct-30. In addition, the initial part of the test set appears quite unstable across multiple channels even though no anomaly is recorded. Therefore we also drop the first 45000 points from the test set. 

\subsection{SKAB}
The Skoltech Anomaly Benchmark\cite{skab} testbed consists of a water circulation system and its control system, along with a data-processing and storage system. Examples of the type of anomalies induced include partial valve closures, connecting shaft imbalance, reduced motor power, cavitation and flow disturbances. Train and test splits are provided by the authors. 

\subsection{MSL and SMAP}
These are expert-labeled datasets from real anomalies encountered during the operation of two spacecraft - Soil Moisture Active Passive (SMAP) satellite and the Mars Science Laboratory (MSL) rover, Curiosity \cite{hundman2018detecting}. Unlike the other datasets, each entity in MSL and SMAP consists of only 1 sensor, while all the other channels are one-hot-encoded commands given to that entity. We use all channels as input to the model, but the model error of only the sensor channel is used for anomaly detection, as done by \cite{hundman2018detecting}.
The total number of variables is 1375 and 1485 for SMAP and MSL respectively, making these much larger than the single-entity datasets in terms of number of variables. The data is however divided into 55 and 27 entities respectively. The authors provide train-test splits so that for the first anomaly encountered in test at time t, the training set is from time t-5 days to t-3 days (if available), and the test set goes from t-3 days to t+2 days. The data is sampled each minute, so the training set is much shorter than other datasets. 

\subsection{SMD}
SMD, or Server Machine Dataset was published by \cite{su2019omni} on their Github repository \url{https://github.com/NetManAIOps/OmniAnomaly}. The data was collected over 5 weeks from a large internet company. It consists of data from 28 entities regularly sampled every minute. The train-test split is 50\% each for train and test, suggested by the authors. An interpretation label is provided for each anomaly which we use for anomaly diagnosis. 

\begin{figure}[h]
    \centering
    \includegraphics[width=\linewidth]{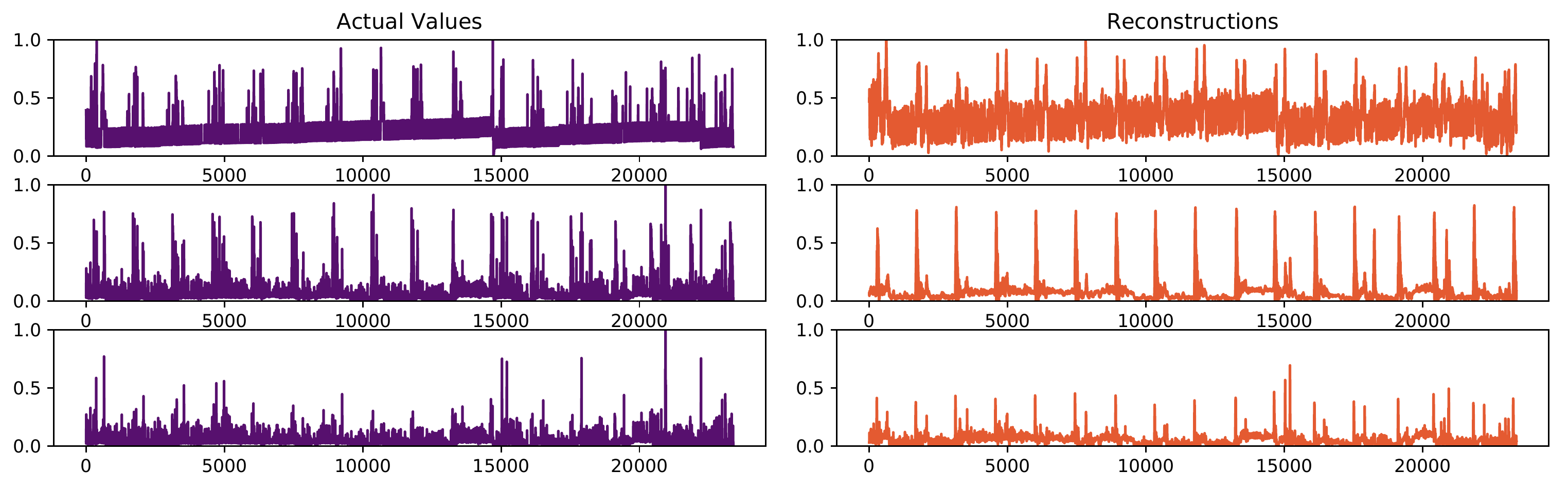}
    \caption{Left: Signals from three channels from the SMD dataset, showing strong correlations across channels and time. Right: Reconstructions of the three channels by Univariate AutoEncoder, accounting for only temporal correlations.}
    \label{fig:sample_ts}
\end{figure}
\clearpage

\section{Additional details about models}

\textbf{Univariate Fully-Connected Auto-Encoder (UAE)}: In this method, we train a separate auto-encoder for each channel. Each encoder is a multi-layer perceptron with $l_w$ nodes in the input to $p$ dimensions in the latent space, with the number of dimensions reducing in powers of 2 in each layer, similar to \cite{audibert2020usad}. The decoder is a mirror image of the encoder, and we use tanh activation. 

\textbf{Fully-Connected Auto-Encoder (FC AE)}: This is similar to UAE, but now we have a single model over all the channels. Thus, the input sample is a flattened subsequence, a vector of size $l_w \times m$. 

\textbf{Long Short Term Memory Auto-Encoder (LSTM AE)}: We use a single LSTM layer for each of the encoder and decoder.  \cite{malhotra2016lstm} used only the first principal component (PC) of the MVTS as input to LSTM-ED, but since this can result in major information loss, we choose the number of PCs corresponding to 90\% explained variance. We set the hidden size to be the same size as the number of PCs.

\section{Hyperparameters and Implementation}
\label{sec:si-hyperparams}
In some algorithms, we set the architecture size as a function of $m$ to accommodate different datasets, eg. the hidden size. All other hyperparameters are kept constant across datasets, obtained by hyperparameter tuning on the SWaT dataset only. All the deep learning methods except OmniAnomaly and OCAN are trained for a maximum of 100 epochs with early stopping using the reconstruction or prediction error on the 25\% held-out validation set and patience = 10. 
\begin{table*}[h]
\centering
\caption{Online code references}
\label{tab:code-refs}
\begin{tabular}{ll}
\hline
Model           & Adapted from \\ \hline
LSTM VAE & \hyperlink{https://github.com/TimyadNyda/Variational-Lstm-Autoencoder}{https://github.com/TimyadNyda/Variational-Lstm-Autoencoder} \\
NASA LSTM & \hyperlink{https://github.com/khundman/telemanom}{https://github.com/khundman/telemanom}\\
DAGMM & \hyperlink{https://github.com/danieltan07/dagmm}{https://github.com/danieltan07/dagmm} \\ 
OmniAnomaly & \hyperlink{https://github.com/NetManAIOps/OmniAnomaly}{https://github.com/NetManAIOps/OmniAnomaly}   \\
MSCRED & \hyperlink{https://github.com/Zhang-Zhi-Jie/Pytorch-MSCRED}{https://github.com/Zhang-Zhi-Jie/Pytorch-MSCRED}, \hyperlink{https://github.com/SKvtun/MSCRED-Pytorch}{https://github.com/SKvtun/MSCRED-Pytorch}\\
OCAN & \hyperlink{https://github.com/PanpanZheng/OCAN}{https://github.com/PanpanZheng/OCAN}\\
BeatGAN & \hyperlink{https://github.com/Vniex/BeatGAN}{https://github.com/Vniex/BeatGAN}\\
\hline
\end{tabular}
\end{table*}
\begin{table*}[!ht]
\centering
\caption{Hyperparameters are tuned using the minimum validation reconstruction error criterion, based only on the SWaT dataset. We tested 50 random hyperparameter configurations for each of TCN AE, LSTM VAE and OC-SVM, and 20 random configurations for FC AE. The chosen hyperparameter is shown in the `Value' column. $m$ is the number of channels in each entity of a dataset. }
\label{tab:hyperparam_tuning}
\begin{tabular}{llll}
\hline
Model           & Hyperparameters & Sampling distribution & Value \\ \hline
FC AE & learning rate         & log-unif($\log(10^{-4})$, $\log(10^{-3})$) & $10^{-4}$\\
                & z-dim                 & int of $\{\frac{m}{5}, \frac{m}{4}, \frac{m}{3}, \frac{m}{2}, \frac{m}{1}, 2m, 3m, 4m, 5m\}$ & $int(\frac{m}{2})$\\
TCN AE & learning rate         & log-unif($\log(10^{-4})$, $\log(10^{-2})$) & $1.5\times 10^{-4}$\\
                & z-dim                 & int $\in [3, 10]$ & 8 \\
                & dropout rate          & unif(0.2, 0.5) & 0.42\\
LSTM VAE        & learning rate         & log-unif($\log(10^{-4})$, $\log(10^{-2})$) & $9.5\times 10^{-3}$\\
                & (z-dim, hidden-dim)                 & $\{(3, 15), (3, 5)\}$ & (3, 15)\\
                & $\lambda_{reg}$       & unif(0,1) & 0.55\\
                & $\lambda_{kulback}$   & unif(0,1) & 0.28\\
OC-SVM          & $\gamma$              & $\left\{\frac{1}{\#features}, \text{uniform}(10^{-5},10^5)\right\}$ & $\frac{1}{\#\text{features}}$\\
                & $\nu$                 & unif(0,0.5) & 0.489\\                      
\hline
\end{tabular}
\end{table*}

\begin{table}[ht]
\centering
\caption{Key hyperparameter values used for each models. $m$ is the number of channels in an entity of the dataset, LR is the learning rate, p refers to the hidden size.
* Learning rate annealing was used as per \cite{su2019omni}}
\label{tab:algo-params-all}
\resizebox{\linewidth}{!}{%
\begin{tabular}{llllp{0.3\linewidth}l}
\toprule
Model                  & Epochs & LR                 & Batch size & Design  & Framework                                                                  \\ \midrule
PCA                    & -      & -                             & -          & $n_{PCA,0.9}$=components for explained variance=0.9   & Scikit-learn                         \\
OC-SVM                 & -      & -                             & -          & $\gamma$=1/m, $\nu$=0.489    &                                                 \\
UAE        & 100    & 0.001                         & 256        & p=5      & Pytorch                                                               \\
FC AE                 & 100    & 0.0001                         & 128        & p=int(m/2)       & Pytorch                                                            \\
LSTM AE               & 100    & 0.001                         & 64         & PCA before model with $n_{PCA, 0.9}$,   hidden-layers=3  & Pytorch                          \\
TCN AE                & 100    & $1.5 \times 10^{-4}$ & 128        & dropout=0.42, p=3, hidden-layers=min(10, int(m/6)), kernel-size=5  & Pytorch               \\
LSTM   VAE             & 100    & $9.5 \times 10^{-3}$                   & 128        & Hidden layers=2, hidden-dim=15, z-dim=3, $\lambda_{reg}$=0.55, $\lambda_{kulback}$=0.28  & Tensorflow\\
BeatGAN & 100 & $10^{-4}$ & 128 & z-dim=10, beta1=0.5 & Pytorch\\
MSCRED & 100 & $10^{-4}$ & 128 & As in \cite{zhang2019deep}. For WADI, PCA before model with $n_{PCA,0.9}$ & Pytorch\\
DAGMM                  & 200    & $10^{-4}$        & 128        & z-dim=m, $GMM_k$=3, $\lambda_{ energy}$=0.1, $\lambda_{cov}$=0.005    & Pytorch        \\
NASA LSTM NPT & 100 & $10^{-3}$ & 64 & LSTM-units=m, LSTM-layers=2, dropout=0.3, NPT params from \cite{hundman2018detecting}    & Keras  \\
NASA LSTM & 100    & $10^{-3}$        & 64         & LSTM units=m, LSTM layers=2, dropout=0.3    & Keras                                       \\
OmniAnomaly     & 20  & 0.001*                 & 50 & All params as in \cite{su2019omni}: z-dim=3, RNN-units=500, dense-units=500, NF-layers=20 & Tensorflow\\
OCAN & 100 & $10^{-4}$ & 128 & As in \cite{zheng2019ocan} & Tensorflow\\
\bottomrule
\end{tabular}
}
\label{table:hyperparam}
\end{table}
\begin{table}
\centering
\caption{Training time for deep learning algorithms in minutes for the single-entity datasets (with early stopping), trained on a single Nvidia GeForce RTX 2080 Ti GPU. UAE, the top performing model in the evaluation, is the third slowest. The slow speed of UAE training is because we trained the channel-wise models sequentially but since each model is independent, it is easily parallelizable. On the other hand, FC AE the second best model, is the fastest to train.}
\label{tab:train-times}
\begin{tabular}{@{}lllll@{}}
\toprule
                        & DAMADICS & SWaT   & WADI   &  \\ \midrule
BeatGAN                 & 9         & 10     & 19     &\\
DAGMM                   & 37    & 45  & 115 &  \\
FC   AE        & 6     & 11  & 2   &  \\
LSTM AE      & 23    & 116 & 142 &  \\
LSTM VAE              & 98    & 93  & 77  &  \\
MSCRED                  & 221         & 510     & 330 & \\
NASA LSTM               & 22    & 25  & 14  &  \\
OCAN                    & 41      & 42    & 121  &\\
OmniAnomaly             & 186   & 165 & 262 &  \\
TCN AE       & 20    & 20  & 38  &  \\
UAE & 65    & 104 & 268 &  \\ \bottomrule
\end{tabular}
\end{table}
\clearpage

\section{Comparison with published results}
\label{sec:compare_published}
\begin{table}[!ht]
\caption{Point-adjusted $F_1$ score for the MSL, SMAP and SMD datasets with the best-f1 threshold, comparing results reported in the literature against UAE Gauss-D and Random Anomaly Detector (discussed in section 6 in the main text).}
\label{tab: f1_pa_compare}
\centering
\begin{tabular}{@{}lrrr@{}}
\toprule
Model                               & MSL    & SMAP   & SMD    \\ \midrule
OmniAnoAlgo (reported by authors)\cite{su2019omni}   & 0.9014 & 0.8535 & 0.9620  \\
OmniAnoAlgo (reproduced by us)     & 0.8459 & 0.8678 & 0.9424 \\ 
USAD (reported by authors)\cite{audibert2020usad} &  0.9109  &  0.8186  & 0.9382  \\
Random Anomaly Detector &  0.8512 & 0.7418   &  0.7585  \\
UAE Gauss-D  & \textbf{0.9204} & \textbf{0.8961}  & \textbf{0.9723} \\
\bottomrule
\end{tabular}
\end{table}
Here we discuss the results in our paper with other published works on the datasets we use.

\textbf{Comparisons on WADI dataset with the point-wise $F_1$ score}: Recently, \cite{audibert2020usad} propose USAD, an adversarially trained auto-encoder model for MVTS anomaly detection, and tested it on 5 of the 6 datasets that we test here. They report a point-wise $F_1$ of 0.2328 with the best-F-score threshold on the WADI dataset. \cite{li2019mad} report $F_1$ score of 0.37 on the WADI dataset with the best-F-score threshold using a Generative Adversarial Network, MAD-GAN. We obtain comparable or better scores in this work, shown in Table \ref{tab: f1_bestf1_gauss-d} using the Gauss-D scoring function and Table \ref{tab: f1_bestf1_gauss-dk} with the Gauss-D-K scoring function. The best overall algorithm, UAE, attains an $F_1$ score of 0.4740 (an improvement of 46.5\% over MAD-GAN) with the Gauss-D-K scoring function, while the best performing algorithm on WADI in this table is LSTM VAE, with an $F_1$ score of 0.5025.

\textbf{Comparison with point-adjusted $F_1$ results}: \cite{su2019omni} and \cite{audibert2020usad} report results on SMAP, MSL and SMD datasets with the point-adjusted $F_1$ score with the OmniAnomaly and USAD algorithms respectively. Based on the experiments discussed in the main text section 6, we do not use this metric to draw comparisons in this paper. However, for the sake of completeness, we show a comparison of the scores of UAE using Gauss-D threshold vs. the results reported by the authors of OmniAnomaly and USAD in Table \ref{tab: f1_pa_compare}. Once again, UAE is the top performing algorithm. 

Aside from the works discussed above, point-wise $F_1$ score with the best-F-score threshold have been published previously on the SWaT dataset. \cite{li2019mad} report a score of 0.77 with the MAD-GAN algorithm, and \cite{audibert2020usad} report a score of 0.79 with the USAD algorithm. However these results are not directly comparable with our results. This is because in our study, we have shortened a long anomaly that biases the results (discussed in section III in the main text) in the SWaT dataset, and as a result our test set is different from these works. We note that some algorithms we tested indeed achieved better $F_1$ score with the best-F-score threshold than the literature methods on SWaT dataset, but these algorithms were not the top-performing by $Fc_1$ score. 

\clearpage

\section{Demonstrations of methods}
\begin{algorithm}
    \SetKwData{Left}{left}\SetKwData{This}{this}\SetKwData{Up}{up}
    \SetKwFunction{Union}{Union}\SetKwFunction{FindCompress}{FindCompress}
    \SetKwInOut{Input}{input}\SetKwInOut{Output}{output}
    \Input{Anomaly-free time-series: $\mathbf{X}_{train} \in \mathbb{R}^{{n_1}\times{m}}$
    \newline
    Test time-series with anomalies: $\mathbf{X}_{test} \in \mathbb{R}^{{n_2}\times{m}}$
    \newline
    Window size $l_w$; step size $l_s$; tail-probability $\epsilon$}
    \Output{Predicted binary anomaly labels $\hat{y_t}$ for each time point in $n_2$
    \newline
    Channel-wise anomaly scores $\mathbf{A}^1$, ..., $\mathbf{A}^m$ each of size $n_2$}
    
    \textbf{Step 1: Train FC AE}
    Hidden size $p$
    \newline
    \For{epoch in max epochs}{
        $Loss\leftarrow 0$
        \newline
        \For{Training sub-sequences $\mathbf{S}_{t, train}\leftarrow \mathbf{X}_{train}[t - l_w + 1, .. ,  t] \in \mathbb{R}^{{l_w} \times{m}}$ with step size $l_s = 10$}{
            Encode sub-sequence to latent space:
            \newline
            $z^1, .. , z^p\leftarrow Encoder(\mathbf{S}_{t, train})$
            \newline
            Reconstruct original sub-sequence:
            \newline
            $\hat{\mathbf{S}}_{t, train}\leftarrow Decoder(z^1, .. , z^p)$
            \newline
            $Loss\leftarrow Loss + rms(\mathbf{S}_{t, train} - \hat{\mathbf{S}}_{t, train})$
        }
        Update parameters of $Encoder$ and $Decoder$ to minimize $Loss$
    }
    \textbf{Step 2: Apply Gauss-S scoring function} 
    \newline
    Obtain reconstruction errors from FC AE on train: 
    $\mathbf{E}_{t, train}^{i} = \mathbf{S}_{t, train}^{i}[t] - \hat{\mathbf{S}}_{t, train}^{i}[t]$,for each channel $i \in m$
    \newline
    Fit $\mathbf{E}^{i}_{train} \sim N(\mu^i, \sigma^i)$ for each channel $i \in m$
    \newline
    Get the reconstruction errors $\mathbf{E}_{t}^{i} = \mathbf{S}_{t, test}^{i}[t] - \hat{\mathbf{S}}_{t, test}^{i}[t]$ on test data
    \newline
    \For{$t\leftarrow 1$  \KwTo $n_2$}{
            Get probability scores for each channel
            \newline
            \For{each channel $i\leftarrow 1$  \KwTo $m$}{
                $\mathbf{A}^{i}_{t}\leftarrow \log(1 - \Phi(\frac{\mathbf{E}_{t}^i-\mu^{i}}{\sigma^i})), \Phi$ is the cdf of $N(0, 1)$
            }
            $\mathbf{a}_{t}\leftarrow -\sum_{i=1}^{m} \mathbf{A}^{i}_{t}$
    }
    \textbf{Step 3:} Apply tail-p threshold
    \newline
    $th_{tail-p}\leftarrow -m\log(\epsilon)$
    \newline
    $\mathbf{prediction}\leftarrow  \mathbf{1}_{(x >= th_{tail-p})}(\mathbf{a})$
\caption{MVTS anomaly detection with an auto-encoder model, Gauss-S scoring and tail-p threshold.}
\label{example-algo}
\end{algorithm}
 
\begin{figure*}[!h]
\centering
\subfloat[Anomaly Detection]
{\includegraphics[width=0.45\textwidth]{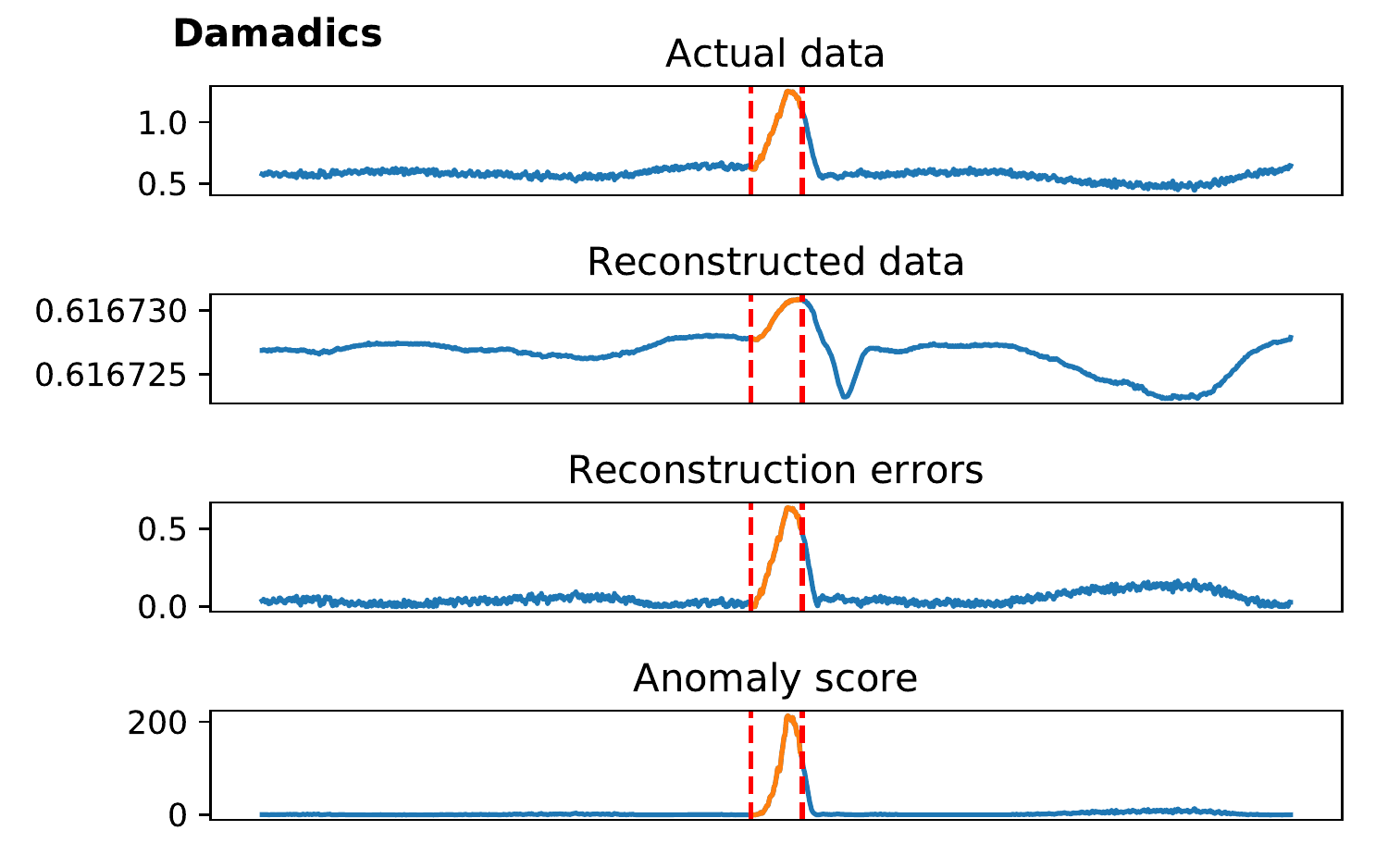}%
\label{Fig:ano-detect-demo}}
\subfloat[Anomaly Diagnosis]{\includegraphics[width=0.45\textwidth]{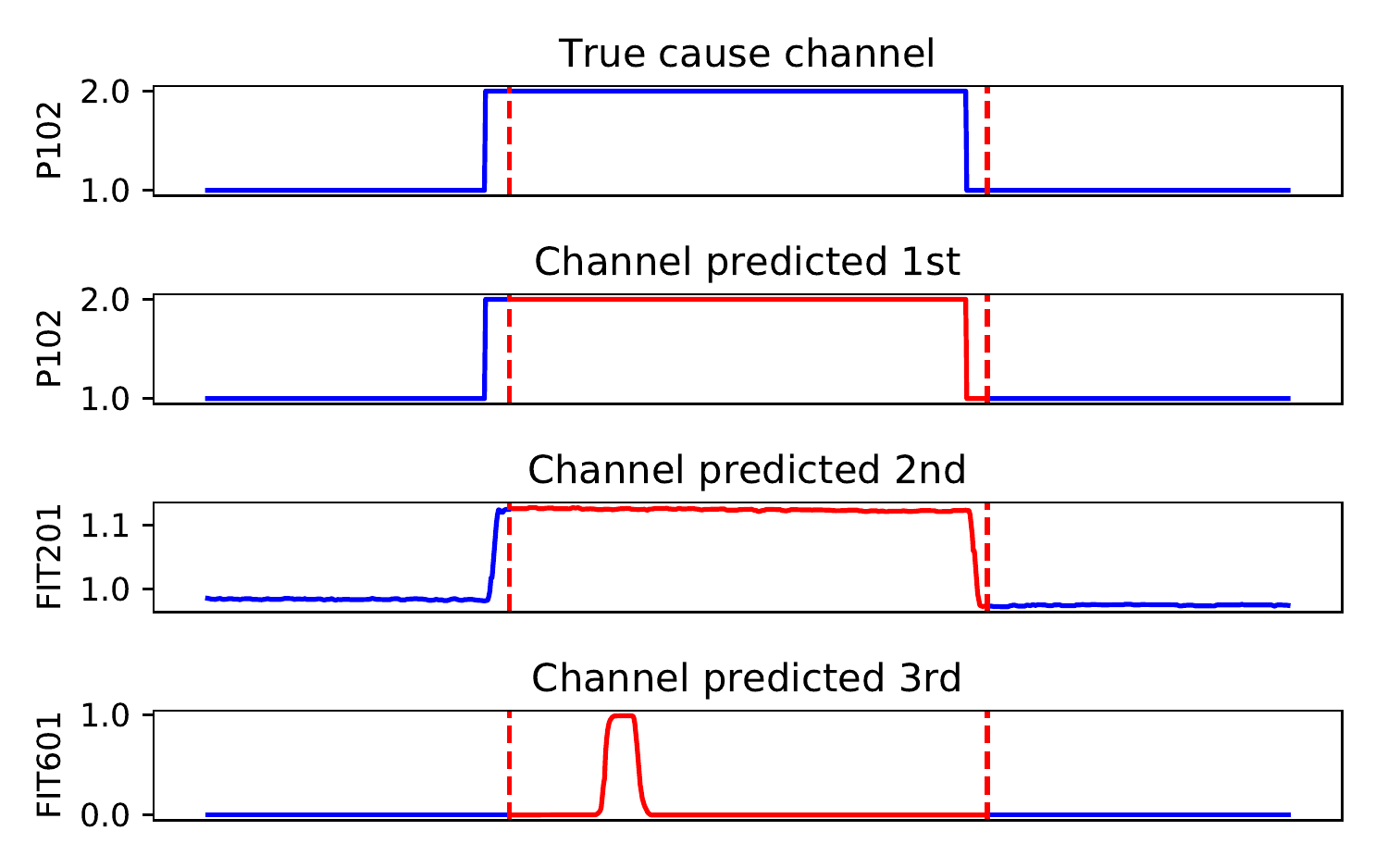}%
\label{Fig:RC}}
\caption{Examples of (a) anomaly detection on DAMADICS dataset and (b) anomaly diagnosis on SWaT dataset using UAE model and Gauss-S scoring function.}
\label{Fig:demo}
\end{figure*}

\begin{figure}[!h]
    \includegraphics[width=1.0\textwidth]{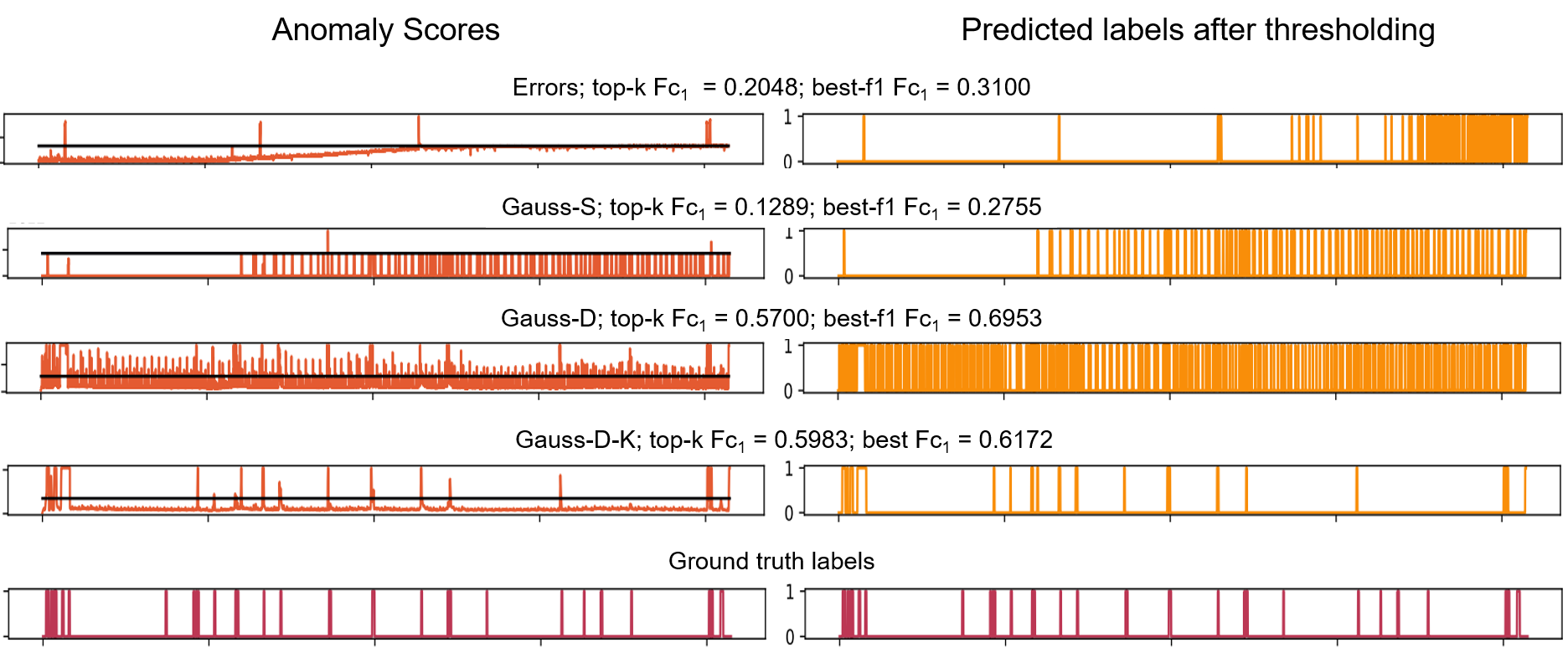}
    \caption{The effect of scoring function on the anomaly detection performance of UAE for the SWaT dataset. Plots on the left show the scoring function after aggregation across channels, and the solid black line is the top-k threshold. Plots on the right show the anomaly labels for each scoring function. Some anomalies stand out with the Error scoring function but others go undetected. Gauss-D has a much better $Fc_1$ score, but it appears noisy. The Gauss-D-K scoring function does smoothing across time and channels with a Guassian kernel, so the scoring function appears much less noisy.
    }
    \label{fig:swat-scoring-funcs}
\end{figure}
\begin{figure}[h]
    \centering
    \includegraphics[width=0.7\linewidth]{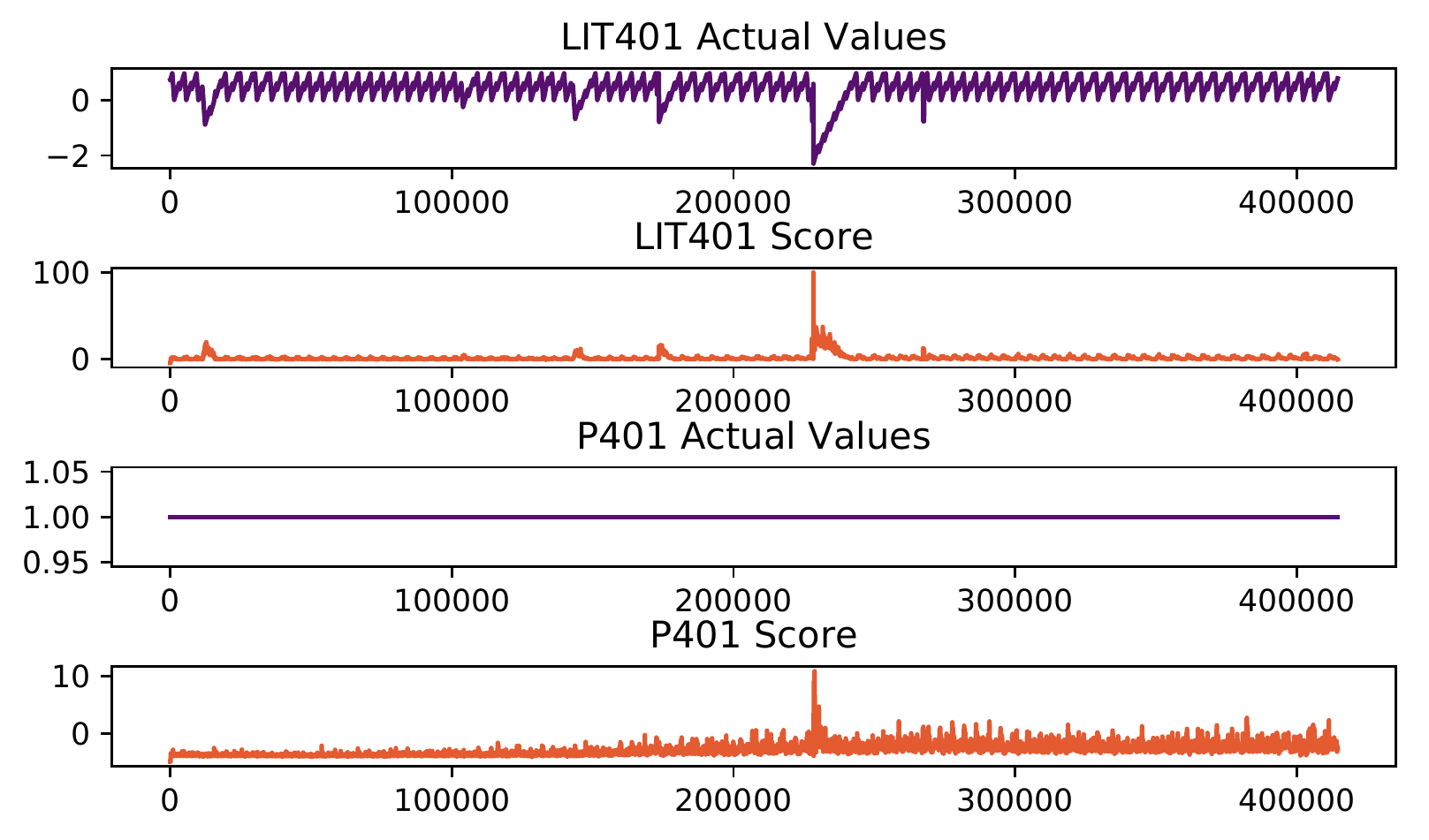}
    \caption{An example showing spurious correlations learnt by OmniAnomaly on SWaT. Channel P401 does not vary at all during test, but OmniAnomaly's score for channel P401 shows peaks corresponding to an anomaly in LIT 401. The signals are shown from the test set. P401 stays constant in both train and test. }
    \label{fig:omni_spurious}
\end{figure}
\clearpage

\section{Statistical Tests}
\label{sec:statistical-tets}
We follow recommendations from  \cite{demvsar2006statistical} for multiple classifier comparisons across multiple datasets. For each comparison, we first use the Friedman test to find whether the overall comparison between k methods on N groups (eg. datasets) is statistically significant. If this comparison is significant, we follow this with post-hoc tests using Hochberg's step-up procedure \cite{hochberg1988sharper} to compare the best method against all other methods. All tests below are conducted at significance level $\alpha = 0.05$. 
\subsection{Effect of scoring functions on anomaly detection performance}
\label{sec:stat-detection-scoring}
The null hypothesis for Friedman test is that all scoring functions perform the same over N=70 combinations of datasets and models, and k=4 scoring functions. The Friedman-statistic is 57.13 which is larger than the critical value resulting in a p-value of $2.4e-12$, thus we reject the null hypothesis. Post-hoc tests using Hochberg's step-up procedure indicate that the difference between the performance of Gauss-D-K vs. all other methods is statistically significant. Furthermore, the difference between the performance of Gauss-D vs. Gauss-S and Error is statistically significant. 
\subsection{Effect of model on anomaly detection performance}
\label{sec:stat-detection-model}
The null hypothesis for Friedman test is that all algorithms in Table IV (main text) perform the same with N=7 datasets and k=13 methods. We find that the Friedman statistic 43.53 is greater than the exact tabled critical value 20.23\cite{lopez2019friedman}, resulting in p-value $1.83e-5$ and reject the null hypothesis. Next we compare UAE against other models using post-hoc tests \cite{hochberg1988sharper} and find that only the comparisons between UAE and the bottom 6 algorithms are statistically significant. 
\clearpage
\section{Additional Results}

\subsection{$Fc_1$ score with top-k threshold}
\label{sec:supp-fc-topk}
\begin{table*}[h]
\setlength{\tabcolsep}{4pt}
\caption{$Fc_1$ score mean and standard deviation over 5 seeds, with the top-k threshold using the chosen hyperparameters. The scoring function is Gauss-D for all algorithms except those denoted with *. The overall mean is a mean over all the datasets. The performance of Raw Signal and PCA models is deterministic.}
\label{tab: fc-top-k-std}
\resizebox{\textwidth}{!}{%
\begin{tabular}{l|rr|rr|rr|rr|rr|rr|rr|p{0.06\textwidth}p{0.06\textwidth}}
\toprule
{} & \multicolumn{2}{c}{DMDS} & \multicolumn{2}{c}{MSL} & \multicolumn{2}{c}{SKAB} & \multicolumn{2}{c}{SMAP} & \multicolumn{2}{c}{SMD} & \multicolumn{2}{c}{SWaT} & \multicolumn{2}{c}{WADI} & Mean & Rank \\
Algo &    Mean &     Std &    Mean &     Std &    Mean &     Std &    Mean &     Std &    Mean &     Std &    Mean &     Std &    Mean & \multicolumn{3}{l}{Std} \\
\midrule
Raw Signal  &  0.4927 &         &  0.2453 &         &  0.5349 &  0.0000 &  0.2707 &         &  0.5151 &         &  0.3796 &         &  0.4094 &         &       0.4068 &      9.3 \\
PCA         &  0.5339 &         &  0.4067 &         &  0.5524 &  0.0000 &  0.3793 &         &  0.5344 &         &  0.5314 &         &  0.3747 &         &       0.4733 &      5.6 \\
UAE         &  \textbf{0.6378} &   0.008 &  \textbf{0.5111} &  0.0085 &  \textbf{0.5550} &  0.0022 &  \textbf{0.4793} &  0.0074 &  0.5501 &  0.0046 &  \textbf{0.5713} &  0.0087 &  0.5105 &    0.01 &       \textbf{0.5450} &     \textbf{ 1.6} \\
FC AE       &  0.6047 &   0.005 &  0.4514 &  0.0048 &  0.5408 &  0.0040 &  0.3788 &  0.0056 &  0.5395 &  0.0064 &  0.4478 &  0.0528 &  0.5639 &  0.0193 &       0.5038 &      4.7 \\
LSTM AE     &  0.5999 &  0.0141 &  0.4481 &  0.0065 &  0.5418 &  0.0054 &  0.4536 &  0.0109 &  0.5271 &  0.0062 &  0.5163 &  0.0148 &  0.4265 &  0.0057 &       0.5019 &      4.7 \\
TCN AE      &  0.5989 &  0.0204 &  0.4354 &  0.0105 &  0.5488 &  0.0041 &  0.3873 &  0.0054 &  \textbf{0.5800} &  0.0037 &  0.4732 &  0.0114 &  0.5126 &  0.0784 &       0.5052 &      3.9 \\
LSTM VAE    &  0.5939 &  0.0085 &  0.3910 &  0.0059 &  0.5439 &  0.0022 &  0.2988 &   0.004 &  0.5427 &  0.0046 &  0.4456 &   0.003 &  \textbf{0.5758} &  0.0143 &       0.4845 &      6.0 \\
BeatGAN     &  0.5391 &  0.1099 &  0.4531 &  0.0075 &  0.5437 &  0.0063 &  0.3732 &  0.0091 &  0.5479 &  0.0099 &  0.4777 &  0.0061 &  0.4908 &  0.0558 &       0.4894 &      5.0 \\
MSCRED      &  0.2906 &  0.0129 &  0.3944 &  0.0045 &  0.5526 &  0.0076 &  0.3724 &  0.0062 &  0.4145 &  0.0057 &  0.4315 &  0.0117 &  0.3253 &  0.0033 &       0.3973 &      8.1 \\
NASA LSTM   &  0.1284 &  0.0074 &  0.4715 &  0.0124 &  0.5339 &  0.0100 &  0.4280 &  0.0077 &  0.3879 &  0.0036 &  0.1398 &  0.0143 &  0.1058 &  0.0449 &       0.3136 &      8.9 \\
DAGMM*       &  0.0000 &       0 &  0.1360 &  0.0188 &  0.0000 &  0.0000 &  0.1681 &  0.0205 &  0.0187 &   0.015 &  0.0000 &       0 &  0.0256 &  0.0573 &       0.0498 &     12.9 \\
OmniAnomaly* &  0.1425 &  0.1189 &  0.4120 &  0.0108 &  0.4561 &  0.0264 &  0.3767 &  0.0094 &  0.5002 &  0.0121 &  0.1466 &  0.0985 &  0.2443 &  0.0202 &       0.3255 &      9.4 \\
OCAN*        &  0.2532 &  0.0925 &  0.3009 &  0.0323 &  0.4369 &  0.0271 &  0.2787 &  0.0177 &  0.4614 &  0.0095 &  0.1547 &  0.1502 &  0.0000 &       0 &       0.2694 &     11.0 \\
\bottomrule
\end{tabular}
}
\end{table*}
\begin{figure}[h]
    \centering
    \includegraphics[width=0.5\linewidth]{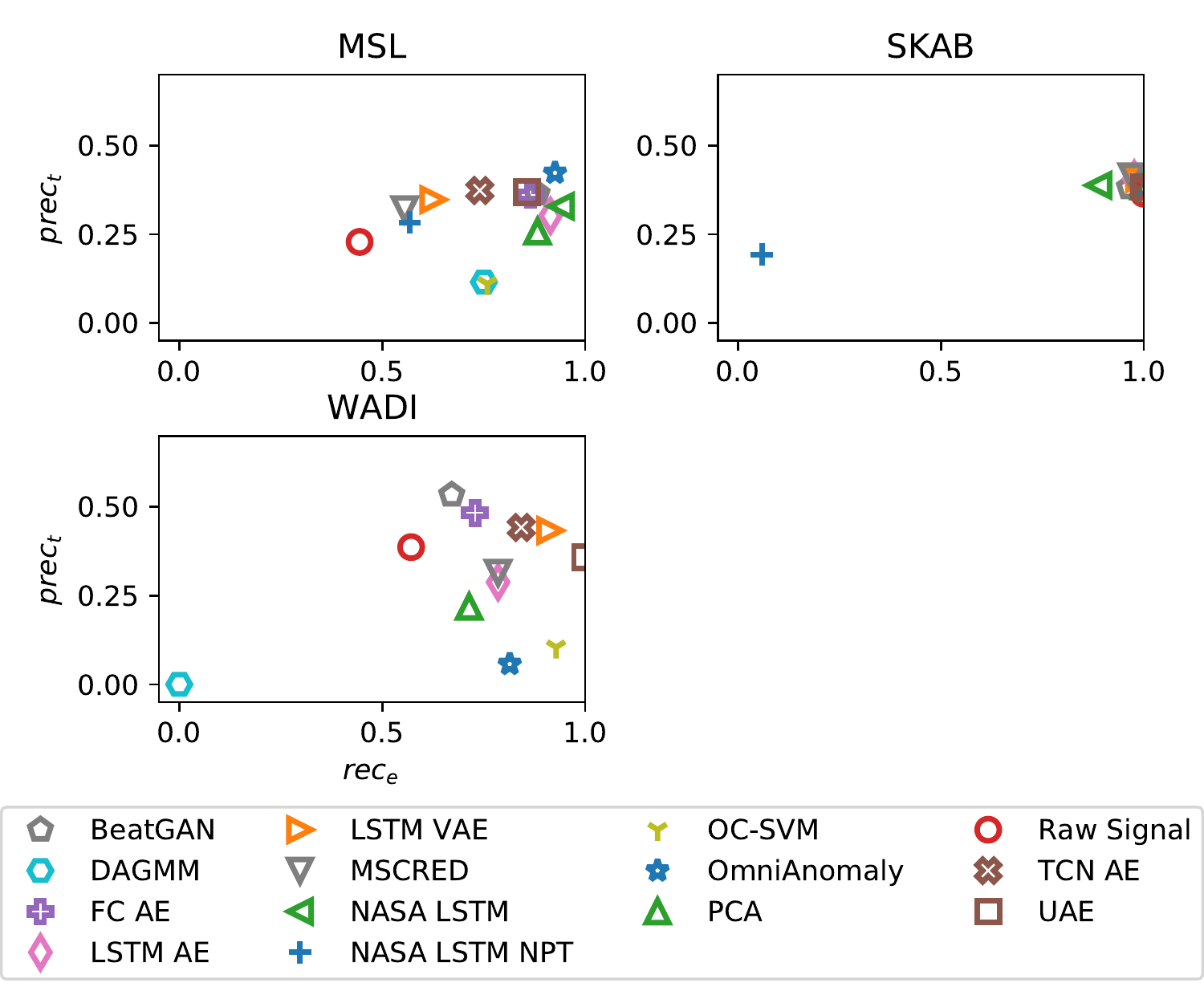}
    \caption{Plots of $prec_t$ vs. $rec_e$ for algorithms with the top-k threshold and scoring functions as in Table IV in main manuscript. See plots for additional datasets in Fig. 4 in main manuscript.}
    \label{fig:supp_prec_rec_topk}
\end{figure}
\begin{table*}[!ht]
\caption{$Fc_1$ score of various models with the \textit{Gauss-D-K} scoring function (except the starred algorithms that specify their own scoring functions) with the \textit{top-k} threshold.}
\label{tab: top-k-gauss-d-k}
\resizebox{\textwidth}{!}{%
\begin{tabular}{l|rr|rr|rr|rr|rr|rr|rr|p{0.06\textwidth}p{0.06\textwidth}}
\toprule
{} & \multicolumn{2}{c}{DMDS} & \multicolumn{2}{c}{MSL} & \multicolumn{2}{c}{SKAB} & \multicolumn{2}{c}{SMAP} & \multicolumn{2}{c}{SMD} & \multicolumn{2}{c}{SWaT} & \multicolumn{2}{c}{WADI} & Mean & Rank \\
Algo &    Mean &     Std &    Mean &     Std &    Mean &     Std &    Mean &     Std &    Mean &     Std &    Mean &     Std &    Mean & \multicolumn{3}{l}{Std} \\
\midrule
Raw Signal  &  0.4693 &         &  0.2974 &         &  0.5405 &  0.0000 &  0.3190 &         &  0.5591 &         &  0.4086 &         &  0.4790 &         &       0.4390 &      7.9 \\
PCA         &  0.5356 &         &  0.5217 &         &  0.5585 &  0.0000 &  0.4585 &         &  0.5557 &         &  0.4811 &         &  0.3068 &         &       0.4883 &      5.4 \\
UAE         &  \textbf{0.6199} &  0.0054 &  0.5193 &   0.021 &  0.5604 &  0.0034 &  0.6116 &  0.0141 &  0.5805 &  0.0041 &  \textbf{0.6105} &  0.0158 &  \textbf{0.5561} &  0.0149 &       \textbf{0.5798} &      \textbf{2.0} \\
FC AE       &  0.6048 &  0.0039 &  0.5060 &  0.0076 &  0.5442 &  0.0042 &  0.5672 &  0.0093 &  0.5651 &  0.0056 &  0.4786 &  0.0475 &  0.5083 &   0.011 &       0.5392 &      4.0 \\
LSTM AE     &  0.6029 &  0.0097 &  0.5346 &  0.0103 &  0.5364 &  0.0202 &  0.5605 &  0.0068 &  0.5381 &  0.0032 &  0.4715 &  0.0092 &  0.3505 &   0.005 &       0.5135 &      5.9 \\
TCN AE      &  0.6035 &  0.0199 &  0.5231 &  0.0057 &  0.5515 &  0.0059 &  0.5615 &  0.0072 &  \textbf{0.6034} &  0.0053 &  0.4353 &  0.0158 &  0.4685 &  0.0854 &       0.5353 &      4.0 \\
LSTM VAE    &  0.5924 &  0.0075 &  0.4623 &  0.0035 &  0.5433 &  0.0042 &  0.4975 &  0.0046 &  0.5784 &  0.0045 &  0.4444 &   0.001 &  0.5289 &  0.0132 &       0.5210 &      5.3 \\
BeatGAN     &  0.5380 &  0.1129 &  0.5329 &  0.0132 &  0.5391 &  0.0095 &  0.5698 &  0.0097 &  0.5550 &  0.0107 &  0.4760 &  0.0266 &  0.4648 &  0.0723 &       0.5251 &      5.3 \\
MSCRED      &  0.2960 &  0.0173 &  0.4096 &   0.007 &  0.5502 &  0.0081 &  0.4009 &  0.0073 &  0.4085 &  0.0045 &  0.3769 &  0.0147 &  0.2905 &  0.0216 &       0.3904 &      8.7 \\
NASA LSTM   &  0.1276 &   0.008 &  \textbf{0.5503} &  0.0098 &  0.5338 &  0.0103 &  \textbf{0.6410} &  0.0131 &  0.3874 &  0.0042 &  0.1348 &  0.0199 &  0.1958 &  0.0446 &       0.3672 &      8.4 \\
DAGMM*       &  0.0000 &       0 &  0.1360 &  0.0188 &  0.0000 &  0.0000 &  0.1681 &  0.0205 &  0.0187 &   0.015 &  0.0000 &       0 &  0.0256 &  0.0573 &       0.0498 &     12.9 \\
OmniAnomaly* &  0.1425 &  0.1189 &  0.4120 &  0.0108 &  0.4561 &  0.0264 &  0.3767 &  0.0094 &  0.5002 &  0.0121 &  0.1466 &  0.0985 &  0.2443 &  0.0202 &       0.3255 &     10.1 \\
OCAN*        &  0.2532 &  0.0925 &  0.3009 &  0.0323 &  0.4369 &  0.0271 &  0.2787 &  0.0177 &  0.4614 &  0.0095 &  0.1547 &  0.1502 &  0.0000 &       0 &       0.2694 &     11.1 \\
\bottomrule
\end{tabular}
}
\end{table*}
\clearpage

\subsection{$Fc_1$ score with best-F-score threshold}
\label{sec:supp-fc-best}
\begin{figure}[!ht]
\centering
    \includegraphics[width=0.5\linewidth]{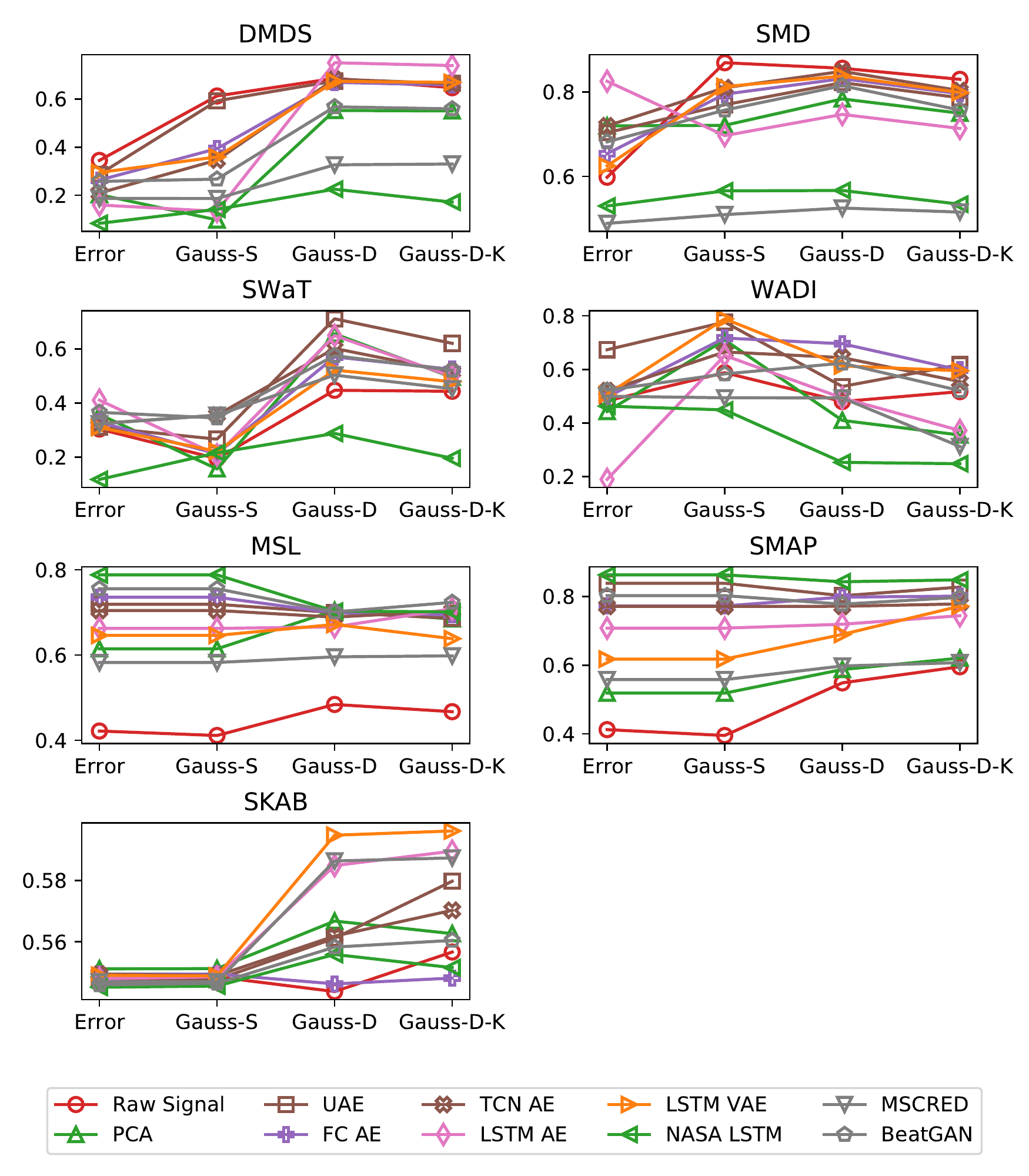}
    \caption{Effect of scoring functions on the $Fc_1$ score using the best-F-score (best-$Fc_1$) threshold.}
    \label{fig:scoring_funcs_bestf1c}
\end{figure}
\begin{table}[!ht]
\caption{$Fc_1$ score of various models with the \textit{Gauss-D} scoring function (except the starred algorithms that specify their own scoring functions) with the \textit{best-F-score} threshold (best $Fc_1$ in this case).}
\label{tab: fc_bestf1}
\resizebox{\textwidth}{!}{%
\begin{tabular}{l|rr|rr|rr|rr|rr|rr|rr|p{0.06\textwidth}p{0.06\textwidth}}
\toprule
{} & \multicolumn{2}{c|}{DMDS} & \multicolumn{2}{c|}{MSL} &
\multicolumn{2}{c|}{SKAB} &
\multicolumn{2}{c|}{SMAP} & \multicolumn{2}{c|}{SMD} & \multicolumn{2}{c|}{SWaT} & \multicolumn{2}{c|}{WADI} & Mean & Rank\\
Algo &    Mean &     Std &    Mean &     Std &    Mean &     Std &    Mean &     Std &    Mean &     Std &    Mean &     Std &    Mean & \multicolumn{3}{l}{Std} \\
\midrule
Raw Signal  &  0.6856 &         &  0.4841 &         &  0.5438 &  0.0000 &  0.5485 &         &  \textbf{0.8568} &         &  0.4469 &         &  0.4801 &         &  0.5780 &      8.0 \\
PCA         &  0.5521 &         &  0.7045 &         &  0.5668 &  0.0000 &  0.5874 &         &  0.7840 &         &  0.6556 &         &  0.4099 &         &  0.6086 &      6.3 \\
UAE         &  0.6744 &  0.0083 &  0.6996 &  0.0104 &  0.5612 &  0.0034 &  0.8024 &  0.0154 &  0.8226 &  0.0075 &  \textbf{0.7112} &  0.0127 &  0.5363 &  0.0185 &  0.6868 &      \textbf{4.0} \\
FC AE       &  0.6687 &  0.0173 &  0.6991 &  0.0078 &  0.5463 &  0.0009 &  0.7984 &  0.0111 &  0.8321 &  0.0026 &  0.5703 &  0.0269 &  \textbf{0.6955} &   0.037 &  \textbf{0.6872} &      5.3 \\
LSTM AE     &  \textbf{0.7504} &  0.0246 &  0.6661 &  0.0096 &  0.5849 &  0.0244 &  0.7193 &  0.0108 &  0.7472 &  0.0066 &  0.6493 &  0.0048 &  0.4934 &   0.047 &  0.6587 &      5.4 \\
TCN AE      &  0.6824 &  0.0208 &  0.6890 &  0.0127 &  0.5619 &  0.0103 &  0.7716 &  0.0189 &  0.8495 &  0.0038 &  0.6016 &  0.0293 &  0.6438 &  0.1075 &  0.6857 &      4.0 \\
LSTM VAE    &  0.6730 &  0.0042 &  0.6718 &   0.025 &  \textbf{0.5948} &  0.0391 &  0.6897 &  0.0207 &  0.8383 &  0.0096 &  0.5215 &  0.0082 &  0.6128 &  0.0089 &  0.6574 &      5.0 \\
BeatGAN     &  0.5673 &  0.1219 &  0.7011 &  0.0189 &  0.5583 &  0.0097 &  0.7778 &  0.0294 &  0.8153 &  0.0078 &  0.5750 &  0.0085 &  0.6230 &  0.0618 &  0.6597 &      5.1 \\
MSCRED      &  0.3262 &  0.0053 &  0.5958 &  0.0053 &  0.5864 &  0.0097 &  0.5977 &  0.0068 &  0.5249 &  0.0083 &  0.5030 &  0.0098 &  0.4935 &  0.0666 &  0.5182 &      8.0 \\
NASA LSTM   &  0.2249 &   0.041 &  0.7030 &  0.0118 &  0.5559 &  0.0158 &  \textbf{0.8431} &  0.0103 &  0.5666 &  0.0047 &  0.2874 &  0.0353 &  0.2535 &  0.0324 &  0.4906 &      8.0 \\
DAGMM*       &  0.0305 &  0.0027 &  0.2601 &  0.0201 &  0.5449 &  0.0049 &  0.3351 &  0.0198 &  0.1066 &  0.0095 &  0.0891 &  0.0004 &  0.1264 &  0.0255 &  0.2132 &     12.7 \\
OmniAnomaly* &  0.2317 &  0.1833 &  \textbf{0.7381} &  0.0188 &  0.5491 &  0.0007 &  0.6782 &   0.019 &  0.7904 &  0.0174 &  0.2836 &  0.1009 &  0.4192 &  0.0396 &  0.5272 &      8.0 \\
OCAN*        &  0.2488 &  0.0988 &  0.5396 &  0.0457 &  0.5482 &  0.0016 &  0.4902 &  0.0173 &  0.6666 &  0.0139 &  0.2541 &  0.1604 &  0.1245 &  0.0117 &  0.4103 &     11.1 \\
\bottomrule
\end{tabular}
}
\end{table}
\clearpage

\subsection{$Fc_1$ score with tail-p threshold}
\label{sec:supp-fc1-unsup}
\begin{table*}[!ht]
\setlength{\tabcolsep}{4pt}
\caption{$Fc_1$ score  mean and standard deviation over 5 seeds, with the tail-p threshold and Gauss-D scoring function, with the following exceptions - * predefined scoring function and tail-p threshold.
$^\dagger$ predefined scoring and threshold. NASA LSTM NPT uses Non-Parametric Threshold. OC-SVM scores are thresholded at 0.5. The value of the threshold, $-\log_{10}(\epsilon) \in \{1:5\}$, and the value that gives the best 
$Fc_1$ is used here. The mean is calculated over all the datasets.
} 
\label{tab: fscore unsup}
\resizebox{\textwidth}{!}{%
\begin{tabular}{lrlrlrrrlrlrlrlrr}
\toprule
dataset & \multicolumn{2}{l}{DMDS} & \multicolumn{2}{l}{MSL} & \multicolumn{2}{l}{SKAB} & \multicolumn{2}{l}{SMAP} & \multicolumn{2}{l}{SMD} & \multicolumn{2}{l}{SWaT} & \multicolumn{2}{l}{WADI} &    Mean & Avg Rank \\
{} &    Mean &     Std &    Mean &     Std &    Mean &     Std &    Mean &     Std &    Mean &     Std &    Mean &     Std &    Mean & \multicolumn{3}{l}{Std} \\
\midrule
Raw Signal    &  0.6192 &         &  0.2618 &         &  0.5340 &  0.0000 &  0.2924 &         &  0.6952 &         &  0.4266 &         &  0.4613 &         &  0.4701 &      8.1 \\
PCA           &  0.2505 &         &  0.3527 &         &  0.5559 &  0.0000 &  0.3638 &         &  0.4915 &         &  0.5979 &         &  0.3349 &         &  0.4210 &      7.0 \\
UAE           &  0.6429 &  0.0139 &  0.4571 &  0.0221 &  0.5545 &  0.0017 &  0.5135 &  0.0181 &  0.5440 &  0.0064 &  0.6467 &   0.011 &  0.5267 &  0.0239 &  0.5551 &      3.1 \\
FC AE         &  0.6630 &  0.0144 &  0.4514 &   0.011 &  0.5429 &  0.0027 &  0.4379 &  0.0123 &  0.5209 &   0.007 &  0.5472 &  0.0307 &  0.5802 &  0.0477 &  0.5348 &      4.9 \\
LSTM AE       &  0.3679 &  0.0162 &  0.4004 &  0.0102 &  0.5725 &  0.0303 &  0.4276 &  0.0077 &  0.4386 &  0.0023 &  0.5903 &  0.0133 &  0.4212 &  0.0058 &  0.4598 &      6.3 \\
TCN AE        &  0.5569 &  0.0414 &  0.4438 &  0.0089 &  0.5484 &  0.0041 &  0.4469 &  0.0117 &  0.5675 &   0.013 &  0.5753 &  0.0346 &  0.5711 &   0.115 &  0.5300 &      4.7 \\
LSTM VAE      &  0.6426 &   0.007 &  0.3917 &  0.0232 &  0.5767 &  0.0376 &  0.3700 &  0.0137 &  0.6175 &  0.0138 &  0.5123 &  0.0144 &  0.5878 &  0.0236 &  0.5284 &      4.7 \\
BeatGAN       &  0.4171 &   0.141 &  0.4656 &  0.0103 &  0.5469 &  0.0190 &  0.3843 &  0.0063 &  0.4475 &  0.0097 &  0.5446 &  0.0208 &  0.5918 &  0.0659 &  0.4854 &      5.4 \\
MSCRED        &  0.1367 &  0.0029 &  0.3629 &  0.0153 &  0.5798 &  0.0105 &  0.3361 &  0.0119 &  0.3085 &  0.0057 &  0.4787 &  0.0088 &  0.4416 &  0.1045 &  0.3778 &      8.1 \\
NASA LSTM     &  0.1130 &  0.0096 &  0.4495 &  0.0134 &  0.5396 &  0.0198 &  0.4410 &  0.0181 &  0.3336 &  0.0058 &  0.2088 &  0.0299 &  0.2252 &  0.0514 &  0.3301 &      8.9 \\
DAGMM*         &  0.0000 &       0 &  0.1800 &  0.0278 &  0.5430 &  0.0063 &  0.2419 &  0.0103 &  0.0000 &       0 &  0.0000 &       0 &  0.0000 &       0 &  0.1378 &     13.0 \\
OmniAnomaly*   &  0.0557 &  0.0012 &  0.5008 &    0.01 &  0.5487 &  0.0006 &  0.4542 &   0.006 &  0.4444 &  0.0108 &  0.1497 &  0.0345 &  0.1072 &  0.0008 &  0.3230 &      7.7 \\
NASA LSTM NPT$^\dagger$ &  0.1440 &   0.003 &  0.3403 &  0.0442 &  0.0902 &  0.0000 &  0.5869 &   0.012 &  0.1957 &  0.0028 &  0.0251 &  0.0057 &  0.1333 &       0 &  0.2165 &     10.3 \\
OC-SVM$^\dagger$         &  0.0337 &       0 &  0.1717 &       0 &  0.5380 &  0.0000 &  0.1757 &       0 &  0.0893 &       0 &  0.0765 &       0 &  0.1876 &       0 &  0.1818 &     12.7 \\
\bottomrule
\end{tabular}
}
\end{table*}
\begin{figure}[!ht]
    \centering
    \includegraphics[width=\textwidth]{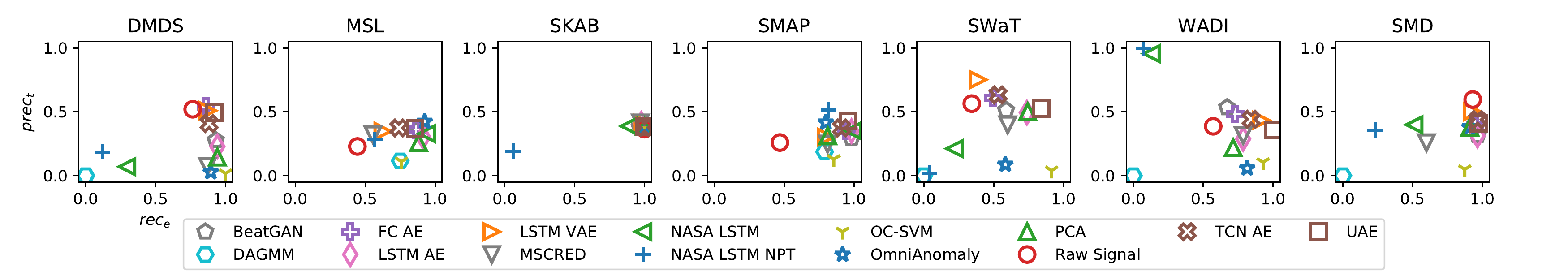}
    \caption{$prec_t$ vs. $rec_e$ for the tail-p threshold for the algorithms shown in Table \ref{tab: fscore unsup}. For NASA LSTM NPT, we use NPT, i.e. non-parametric threshold, and for OC-SVM we threshold at 0.5, instead of tai-p threshold. }
    \label{fig:prec_rec_unsup}
\end{figure}
\clearpage

\subsection{Point-wise $F_1$ score}
\label{sec:supp-f1}
\begin{table}[!ht]
\centering
\caption{Point-wise $F_1$ score (i.e., the $F_1$ score) of various models with the \textit{Gauss-D} scoring function (except the starred algorithms that specify their own scoring functions) with the \textit{best-f1} threshold.}
\label{tab: f1_bestf1_gauss-d}
\begin{tabular}{lrrrrrrrrr}
\toprule
Model &    DMDS &     MSL &    SKAB &    SMAP &     SMD &    SWaT &    WADI &    Mean &  Avg Rank \\
\midrule
Raw Signal  &  0.4015 &  0.3059 &  0.5369 &  0.3009 &  0.4106 &  0.3068 &  0.3634 &  0.3751 &       9.6 \\
PCA         &  0.3958 &  0.3982 &  0.5377 &  0.3334 &  0.4517 &  0.3954 &  0.2846 &  0.3995 &       6.6 \\
UAE         &  \textbf{0.5165} &  \textbf{0.4454} &  0.5375 &  \textbf{0.3937} &  0.4351 &  \textbf{0.4544} &  0.3503 &  \textbf{0.4476} &       \textbf{3.4} \\
FC AE       &  0.4902 &  0.4377 &  0.5375 &  0.3692 &  0.4429 &  0.3545 &  0.4056 &  0.4339 &       4.5 \\
LSTM AE     &  0.4844 &  0.4088 &  0.5372 &  0.3752 &  0.4295 &  0.3956 &  0.3319 &  0.4232 &       6.1 \\
TCN AE      &  0.4654 &  0.4366 &  0.5374 &  0.3729 &  \textbf{0.4826} &  0.3931 &  0.3805 &  0.4384 &       4.1 \\
LSTM VAE    &  0.4724 &  0.4001 &  0.5374 &  0.2998 &  0.4326 &  0.3136 &  \textbf{0.4081} &  0.4091 &       7.1 \\
BeatGAN     &  0.4023 &  0.4193 &  0.5376 &  0.3413 &  0.4529 &  0.3885 &  0.3629 &  0.4150 &       5.3 \\
MSCRED      &  0.2624 &  0.4007 &  0.5411 &  0.3468 &  0.4661 &  0.3558 &  0.2316 &  0.3721 &       6.3 \\
NASA LSTM   &  0.0971 &  0.4104 &  0.5373 &  0.3838 &  0.4228 &  0.1225 &  0.1284 &  0.3003 &       9.1 \\
DAGMM*       &  0.0305 &  0.1938 &  0.5372 &  0.1964 &  0.0662 &  0.0592 &  0.0968 &  0.1686 &      12.8 \\
OmniAnomaly* &  0.1321 &  0.4277 &  0.5400 &  0.3699 &  0.4288 &  0.1726 &  0.2659 &  0.3339 &       7.4 \\
OCAN*        &  0.2365 &  0.3155 &  \textbf{0.5439} &  0.2656 &  0.4437 &  0.1307 &  0.1236 &  0.2942 &       8.9 \\
\bottomrule
\end{tabular}
\end{table}

\begin{table}[!ht]
\centering
\caption{Point-wise $F_1$ score (i.e., the $F_1$ score) of various models with the \textit{Gauss-D-K} scoring function (except the starred algorithms that specify their own scoring functions) with the \textit{best-f1} threshold.}
\label{tab: f1_bestf1_gauss-dk}
\begin{tabular}{lrrrrrrrrr}
\toprule
Model &    DMDS &     MSL &    SKAB &    SMAP &     SMD &    SWaT &    WADI &    Mean &  Avg Rank \\
\midrule
Raw Signal  &  0.4119 &  0.3323 &  0.5369 &  0.3593 &  0.4583 &  0.3760 &  0.4132 &  0.4126 &       9.0 \\
PCA         &  0.4090 &  0.5323 &  0.5376 &  0.4481 &  0.4860 &  0.4453 &  0.3054 &  0.4520 &       6.4 \\
UAE         &  0.5205 &  0.5402 &  0.5382 &  0.5755 &  0.4732 &  0.5799 &  0.4740 &  \textbf{0.5288} &       \textbf{2.9} \\
FC AE       &  0.4942 &  0.5205 &  0.5387 &  0.5421 &  0.4828 &  0.4506 &  0.4902 &  0.5027 &       3.4 \\
LSTM AE     &  0.4923 &  0.5355 &  0.5372 &  0.5317 &  0.4551 &  0.4480 &  0.3247 &  0.4749 &       6.2 \\
TCN AE      &  0.4714 &  0.5493 &  0.5377 &  0.5453 &  0.5195 &  0.4256 &  0.4348 &  0.4977 &       4.1 \\
LSTM VAE    &  0.4834 &  0.4897 &  0.5377 &  0.4875 &  0.4673 &  0.4158 &  0.5024 &  0.4834 &       5.9 \\
BeatGAN     &  0.4055 &  0.5373 &  0.5378 &  0.5254 &  0.4826 &  0.4416 &  0.4155 &  0.4780 &       5.6 \\
MSCRED      &  0.2650 &  0.4479 &  0.5412 &  0.3852 &  0.4671 &  0.3652 &  0.2664 &  0.3911 &       7.7 \\
NASA LSTM   &  0.0998 &  0.5481 &  0.5381 &  0.5933 &  0.4412 &  0.1309 &  0.2031 &  0.3649 &       7.7 \\
DAGMM*       &  0.0305 &  0.1938 &  0.5372 &  0.1964 &  0.0662 &  0.0592 &  0.0968 &  0.1686 &      12.8 \\
OmniAnomaly* &  0.1321 &  0.4277 &  0.5400 &  0.3699 &  0.4288 &  0.1726 &  0.2659 &  0.3339 &       9.4 \\
OCAN*        &  0.2365 &  0.3155 &  0.5439 &  0.2656 &  0.4437 &  0.1307 &  0.1236 &  0.2942 &       9.9 \\
\bottomrule
\end{tabular}
\end{table}
\clearpage

\subsection{AU-ROC score}
\label{sec:supp-auroc}
\begin{table}[!ht]
\centering
\caption{Area under the receiver operator characteristic curve (AU-ROC) of various models with the \textit{Gauss-D} scoring function (except the starred algorithms that specify their own scoring functions).}
\label{tab: auroc-gauss-d}
\begin{tabular}{lrrrrrrrrr}
\toprule
Model &    DMDS &     MSL &    SKAB &    SMAP &     SMD &    SWaT &    WADI &    Mean &  Avg Rank \\
\midrule
Raw Signal  &  0.7073 &  0.4379 &  0.4855 &  0.4953 &  0.7475 &  0.7273 &  0.7733 &  0.6249 &      10.0 \\
PCA         &  0.7413 &  0.6175 &  0.5037 &  0.5684 &  0.7900 &  0.8249 &  0.6583 &  0.6720 &       7.3 \\
UAE         &  0.8852 &  \textbf{0.6712} &  0.5088 &  0.6256 &  \textbf{0.8298} &  \textbf{0.8279} &  0.7405 &  0.7270 &       \textbf{2.6} \\
FC AE       &  \textbf{0.9280} &  0.6602 &  0.4914 &  0.6272 &  0.8028 &  0.8085 &  \textbf{0.7912} &  \textbf{0.7299} &       4.0 \\
LSTM AE     &  0.8404 &  0.6205 &  0.5040 &  0.6137 &  0.8196 &  0.8164 &  0.6808 &  0.6993 &       5.4 \\
TCN AE      &  0.8109 &  0.6184 &  0.5002 &  0.6262 &  0.8282 &  0.7596 &  0.7109 &  0.6935 &       5.5 \\
LSTM VAE    &  0.9146 &  0.5986 &  0.5002 &  0.5628 &  0.7587 &  0.7570 &  0.7892 &  0.6973 &       7.2 \\
BeatGAN     &  0.7960 &  0.6517 &  0.4941 &  0.5993 &  0.8269 &  0.7889 &  0.7086 &  0.6951 &       6.1 \\
MSCRED      &  0.6866 &  0.6436 &  \textbf{0.5270} &  0.6020 &  0.8157 &  0.7069 &  0.6951 &  0.6681 &       6.4 \\
NASA LSTM   &  0.6743 &  0.6580 &  0.4885 &  0.6406 &  0.7538 &  0.5489 &  0.4633 &  0.6039 &       9.1 \\
DAGMM*       &  0.4287 &  0.4761 &  0.5080 &  0.5459 &  0.4767 &  0.4663 &  0.5025 &  0.4863 &      11.1 \\
OmniAnomaly* &  0.6804 &  0.6523 &  0.5075 &  \textbf{0.6588} &  0.7972 &  0.5900 &  0.5275 &  0.6305 &       6.9 \\
OCAN*        &  0.6948 &  0.5803 &  0.5192 &  0.5656 &  0.7736 &  0.5327 &  0.4708 &  0.5910 &       9.3 \\
\bottomrule
\end{tabular}
\end{table}

\begin{table}[h]
\centering
\caption{Area under the receiver operator characteristic curve (AU-ROC) of various models with the \textit{Gauss-D-K} scoring function (except the starred algorithms that specify their own scoring functions).}
\label{tab: auroc-gauss-d-k}
\begin{tabular}{lrrrrrrrrr}
\toprule
Model &    DMDS &     MSL &    SKAB &    SMAP &     SMD &    SWaT &    WADI &    Mean &  Avg Rank \\
\midrule
Raw Signal  &  0.7159 &  0.4815 &  0.4814 &  0.4878 &  0.7917 &  0.7618 &  0.7704 &  0.6415 &       9.7 \\
PCA         &  0.7486 &  0.7448 &  0.4983 &  0.6483 &  0.8275 &  0.8440 &  0.7084 &  0.7171 &       6.4 \\
UAE         &  0.8890 &  \textbf{0.7679} &  0.5079 &  \textbf{0.7643} &  0.8599 &  \textbf{0.8637} &  \textbf{0.8178} &  \textbf{0.7815} &       \textbf{1.9} \\
FC AE       &  \textbf{0.9317} &  0.7594 &  0.4874 &  0.7396 &  0.8375 &  0.8249 &  0.8166 &  0.7710 &       4.3 \\
LSTM AE     &  0.8421 &  0.7548 &  0.4984 &  0.7306 &  0.8410 &  0.8350 &  0.7273 &  0.7470 &       4.7 \\
TCN AE      &  0.8117 &  0.7160 &  0.4977 &  0.7239 &  \textbf{0.8601} &  0.7696 &  0.7514 &  0.7329 &       5.4 \\
LSTM VAE    &  0.9186 &  0.6812 &  0.4967 &  0.6479 &  0.7961 &  0.7766 &  0.8116 &  0.7327 &       6.6 \\
BeatGAN     &  0.7967 &  0.7591 &  0.4886 &  0.7109 &  0.8523 &  0.8066 &  0.7381 &  0.7360 &       5.7 \\
MSCRED      &  0.6898 &  0.6723 &  0.5267 &  0.6374 &  0.8171 &  0.7212 &  0.7233 &  0.6840 &       7.7 \\
NASA LSTM   &  0.6830 &  0.7670 &  0.4876 &  0.7577 &  0.7690 &  0.5663 &  0.5091 &  0.6485 &       8.6 \\
DAGMM*       &  0.4287 &  0.4761 &  0.5080 &  0.5459 &  0.4767 &  0.4663 &  0.5025 &  0.4863 &      11.3 \\
OmniAnomaly* &  0.6804 &  0.6523 &  0.5075 &  0.6588 &  0.7972 &  0.5900 &  0.5275 &  0.6305 &       8.9 \\
OCAN*        &  0.6948 &  0.5803 &  \textbf{0.5192} &  0.5656 &  0.7736 &  0.5327 &  0.4708 &  0.5910 &       9.9 \\
\bottomrule
\end{tabular}
\end{table}

\begin{figure}[!h]
\centering
    \includegraphics[width=0.5\linewidth]{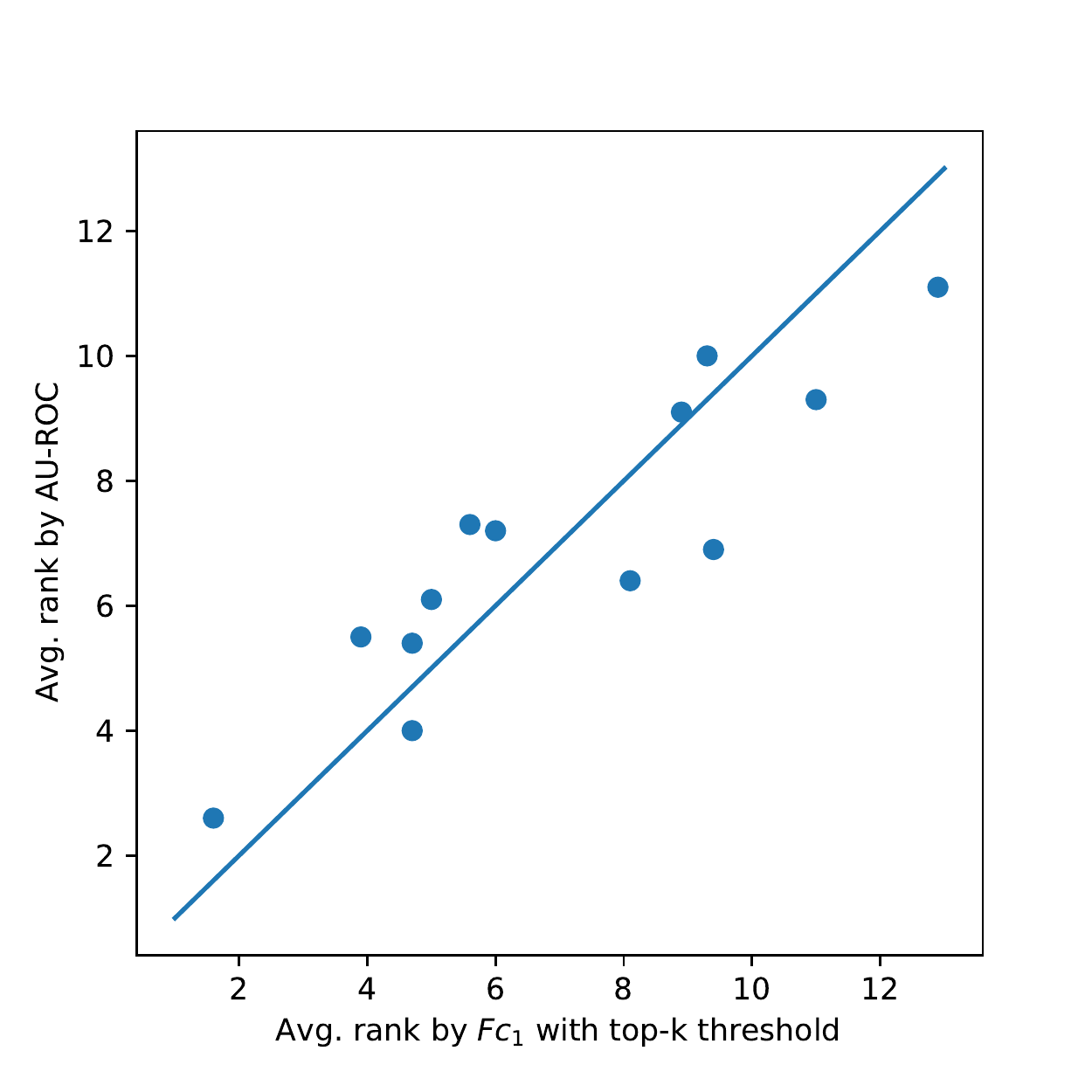}
    \caption{Comparison of average ranks of algorithms from $Fc_1$ score (Table \ref{tab: fc-top-k-std}) vs. AU-ROC score (Table \ref{tab: auroc-gauss-d}). The two metrics are generally in agreement, as evidenced by the closeness of scatter points to the x=y line.}
\end{figure}
\clearpage

\subsection{AU-PRC score (Average Precision)}
\label{sec:supp-auprc}
\begin{table}[!ht]
\centering
\caption{Area under the precision recall curve (AU-PRC), calculated through average precision (AP), of various models with the \textit{Gauss-D} scoring function (except the starred algorithms that specify their own scoring functions).}
\label{tab: auprc-gauss-d}
\begin{tabular}{lrrrrrrrrr}
\toprule
Model &    DMDS &     MSL &    SKAB &    SMAP &     SMD &    SWaT &    WADI &    Mean &  Avg Rank \\
\midrule
Raw Signal  &  0.2946 &  0.2092 &  0.3549 &  0.2121 &  0.3492 &  0.2348 &  0.2525 &  0.2725 &      10.0 \\
PCA         &  0.3144 &  0.2659 &  0.3694 &  0.2618 &  0.3859 &  0.3784 &  0.1520 &  0.3040 &       7.0 \\
UAE         &  0.4134 &  0.3459 &  0.3699 &  0.3220 &  0.3835 &  0.3829 &  0.2186 &  0.3480 &       3.3 \\
FC AE       &  0.4147 &  0.3314 &  0.3535 &  0.2819 &  0.3932 &  0.3298 &  0.3125 &  0.3453 &       4.9 \\
LSTM AE     &  0.3972 &  0.3003 &  0.3683 &  0.2978 &  0.3644 &  0.3704 &  0.1756 &  0.3249 &       6.3 \\
TCN AE      &  0.3776 &  0.3228 &  0.3654 &  0.2965 &  0.4285 &  0.3425 &  0.2761 &  0.3442 &       4.6 \\
LSTM VAE    &  0.3845 &  0.3216 &  0.3666 &  0.2337 &  0.3812 &  0.2843 &  0.2886 &  0.3229 &       6.4 \\
BeatGAN     &  0.3273 &  0.3171 &  0.3628 &  0.2592 &  0.4034 &  0.3388 &  0.2780 &  0.3267 &       6.1 \\
MSCRED      &  0.1298 &  0.3182 &  0.3870 &  0.2624 &  0.4104 &  0.3208 &  0.1521 &  0.2830 &       6.0 \\
NASA LSTM   &  0.0569 &  0.3138 &  0.3616 &  0.3052 &  0.3599 &  0.1011 &  0.0802 &  0.2255 &       9.7 \\
DAGMM*       &  0.0129 &  0.1274 &  0.3709 &  0.1808 &  0.0441 &  0.0434 &  0.0597 &  0.1199 &      11.6 \\
OmniAnomaly* &  0.0844 &  0.3299 &  0.3724 &  0.3090 &  0.3802 &  0.1118 &  0.2081 &  0.2565 &       6.6 \\
OCAN*        &  0.2067 &  0.2224 &  0.3756 &  0.2232 &  0.3958 &  0.1174 &  0.0513 &  0.2275 &       8.6 \\
\bottomrule
\end{tabular}
\end{table}

\begin{table}[h]
\centering
\caption{Area under the precision recall curve (AU-PRC), calculated through average precision (AP), of various models with the \textit{Gauss-D-K} scoring function (except the starred algorithms that specify their own scoring functions).}
\label{tab: auprc-gauss-d-k}
\begin{tabular}{lrrrrrrrrr}
\toprule
Model &    DMDS &     MSL &    SKAB &    SMAP &     SMD &    SWaT &    WADI &    Mean &  Avg Rank \\
\midrule
Raw Signal  &  0.2968 &  0.2528 &  0.3541 &  0.2874 &  0.4060 &  0.2731 &  0.3519 &  0.3174 &       9.1 \\
PCA         &  0.3167 &  0.4582 &  0.3669 &  0.3855 &  0.4358 &  0.3943 &  0.1665 &  0.3606 &       6.0 \\
UAE         &  0.4130 &  0.4543 &  0.3710 &  0.5323 &  0.4334 &  0.4608 &  0.3814 &  0.4352 &       3.3 \\
FC AE       &  0.4150 &  0.4422 &  0.3502 &  0.4871 &  0.4451 &  0.3900 &  0.4315 &  0.4230 &       4.6 \\
LSTM AE     &  0.3976 &  0.4568 &  0.3673 &  0.4773 &  0.4037 &  0.3864 &  0.2395 &  0.3898 &       5.7 \\
TCN AE      &  0.3778 &  0.4753 &  0.3648 &  0.5009 &  0.4802 &  0.3513 &  0.3439 &  0.4135 &       4.3 \\
LSTM VAE    &  0.3822 &  0.4226 &  0.3646 &  0.4318 &  0.4278 &  0.3411 &  0.4073 &  0.3968 &       6.1 \\
BeatGAN     &  0.3268 &  0.4595 &  0.3592 &  0.4710 &  0.4478 &  0.3604 &  0.3577 &  0.3975 &       5.1 \\
MSCRED      &  0.1314 &  0.3779 &  0.3874 &  0.3288 &  0.4115 &  0.3063 &  0.2514 &  0.3135 &       7.3 \\
NASA LSTM   &  0.0585 &  0.4682 &  0.3587 &  0.5573 &  0.3839 &  0.1076 &  0.1538 &  0.2983 &       8.6 \\
DAGMM*       &  0.0129 &  0.1274 &  0.3709 &  0.1808 &  0.0441 &  0.0434 &  0.0597 &  0.1199 &      11.7 \\
OmniAnomaly* &  0.0844 &  0.3299 &  0.3724 &  0.3090 &  0.3802 &  0.1118 &  0.2081 &  0.2565 &       9.4 \\
OCAN*        &  0.2067 &  0.2224 &  0.3756 &  0.2232 &  0.3958 &  0.1174 &  0.0513 &  0.2275 &       9.7 \\
\bottomrule
\end{tabular}
\end{table}

\clearpage

\subsection{Root cause results}
\label{sec:supp-rc}
\begin{table*}[!ht]
\centering
\caption{RC-Top3-all metric for all datasets with root cause labels using the Gauss-D scoring function except starred.}
\label{tab: rc_top3_all_std}
\begin{tabular}{l|rr|rr|rr|rr|p{0.1\textwidth}p{0.1\textwidth}}
\toprule
{} & \multicolumn{2}{c|}{DMDS} & \multicolumn{2}{c|}{SMD} & \multicolumn{2}{c|}{SWaT} & \multicolumn{2}{c|}{WADI} & Overall mean & Avg Rank\\
Algo &     Mean &    Std &    Mean &    Std &    Mean &    Std &    Mean & 
Std \\

\midrule
Raw Signal            &   0.7059 &   &  \textbf{0.9635} &   &  0.3143 &   &  0.5000 &   &       0.6209 & 6.5\\
PCA                   &   0.8235 &   &  0.7831 &   &  0.4857 &   &  0.5000 &   &       0.6481 & 7.2\\
UAE &   \textbf{0.9647} &  0.032 &  0.9498 &  0.006 &  \textbf{0.6286} &  0.020 &  \textbf{0.5428} &  0.039 &       \textbf{0.7715} & \textbf{1.8}\\
FC AE        &   0.9412 &  0.0000 &  0.9522 &  0.0053 &  0.6000 &  0.0286 &  0.4428 &  0.0783 &       0.7341 & 3.8 \\
LSTM AE      &   0.8117 &  0.0263 &  0.9360 &  0.0032 &  0.5771 &  0.0128 &  0.5143 &  0.0319 &       0.7098 & 5.2 \\
TCN AE       &   0.9177 &  0.0526 &  0.9448 &  0.0064 &  0.5143 &  0.0452 &  0.5286 &  0.1083 &       0.7263 & 4.5 \\
LSTM VAE              &   0.9177 &  0.0322 &  0.9501 &  0.0019 &  0.4571 &  0.0000 &  0.5000 &  0.0000 &       0.7062 & 4.8\\
BeatGAN     &  0.9176 &  0.0526 &  0.9424 &  0.0089 &  0.5543 &  0.0256 &  0.4571 &  0.0391 &  0.7178 &      5.8 \\
MSCRED      &  0.7765 &  0.0263 &  0.8526 &  0.0101 &  0.5200 &  0.0424 &  0.0714 &       0 &  0.5551 &      8.2 \\
OmniAnomaly*           &   0.9177 &  0.0671 &  0.9272 &  0.0067 &  0.3772 &  0.0619 &  0.4857 &  0.0319 &       0.6770 & 7.2 \\
\bottomrule
\end{tabular}
\end{table*}

\begin{figure}[!ht]
\centering
    \includegraphics[width=0.5\linewidth]{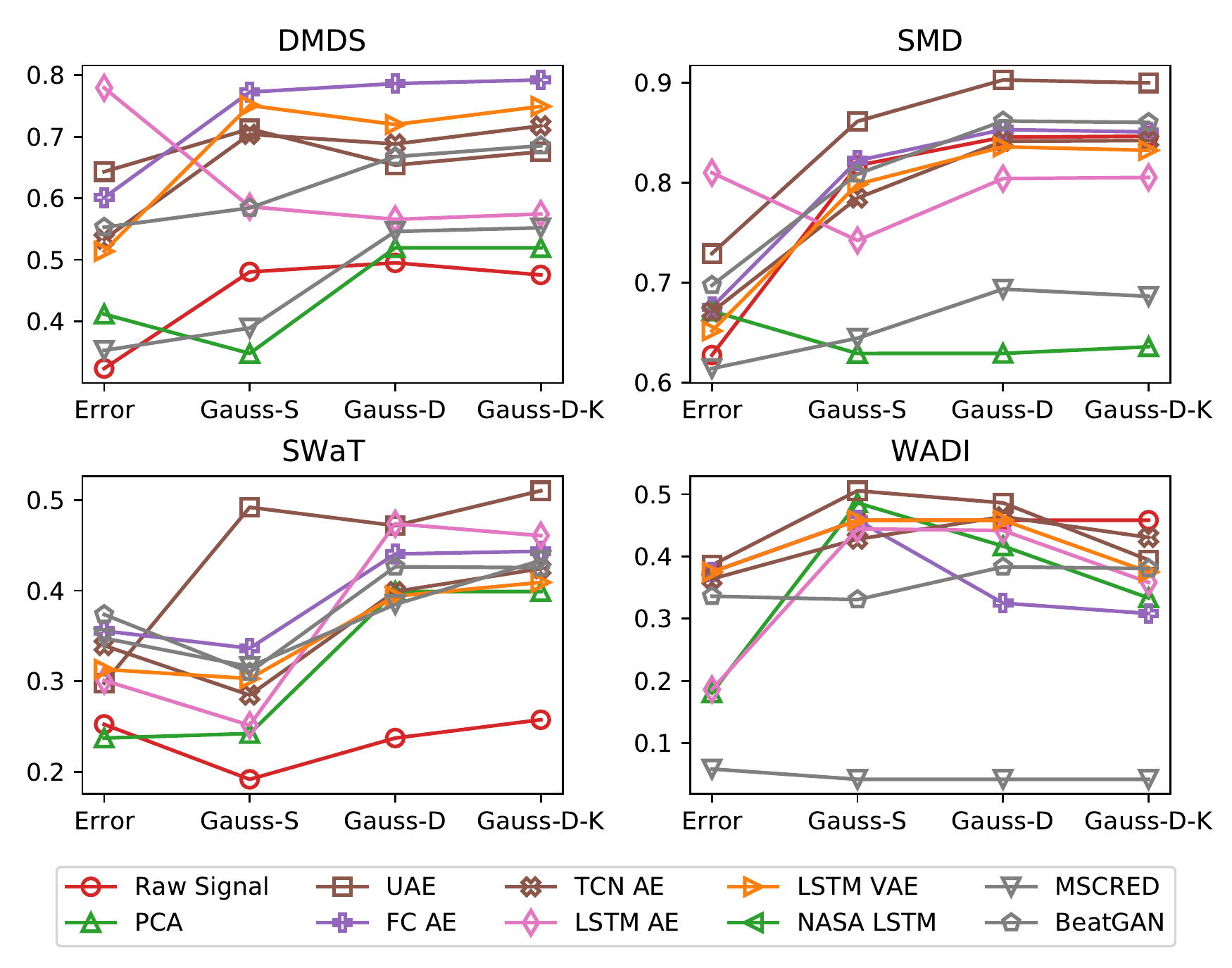}
    \caption{Effect of scoring functions on the HitRate@150 performance for various algorithms and datasets.}
    \label{fig:scoring_funcs_hr150}
\end{figure}
\begin{table*}[!ht]
\centering
\setlength{\tabcolsep}{4pt}
\caption{Hitrate@150 for independent anomaly diagnosis with the Gauss-D scoring function except starred.}
\label{tab:hr_150_all}
\begin{tabular}{l|rr|rr|rr|rr|p{0.1\textwidth}p{0.1\textwidth}}
\toprule
{} & \multicolumn{2}{c|}{DMDS} & \multicolumn{2}{c|}{SMD} & \multicolumn{2}{c|}{SWaT} & \multicolumn{2}{c|}{WADI} & Overall mean & Avg Rank\\
Algo &     Mean &    Std &    Mean &    Std &    Mean &    Std &    Mean & Std \\
\midrule
Raw Signal            &   0.4951 &  0.0000 &  0.8453 &  0.0000 &  0.2374 &  0.0000 &  0.4583 &  0.0000 &       0.5090 & 6.6\\
PCA                   &   0.5196 &  0.0000 &  0.6291 &  0.0000 &  0.3990 &  0.0000 &  0.4167 &  0.0000 &       0.4911 & 7.6 \\
UAE &   0.6540 &  0.011 &  \textbf{0.9026} &  0.003 &  0.4717 &  0.013 &  0\textbf{.4861} &  0.031 &       \textbf{0.6286} & \textbf{2.2}\\
FC AE        &   \textbf{0.7863} &  0.0096 &  0.8529 &  0.0025 &  0.4404 &  0.0083 &  0.3250 &  0.0745 &       0.6012 & 4.0\\
LSTM AE      &   0.5657 &  0.0281 &  0.8039 &  0.0089 &  \textbf{0.4737} &  0.0206 &  0.4417 &  0.0410 &       0.5712 & 5.2\\
TCN AE       &   0.6883 &  0.0356 &  0.8413 &  0.0031 &  0.3990 &  0.0279 &  0.4639 &  0.0999 &       0.5981 & 3.9\\
LSTM VAE              &   0.7226 &  0.0172 &  0.8357 &  0.0009 &  0.3939 &  0.0000 &  0.4583 &  0.0000 &       0.6026 & 4.6 \\
BeatGAN     &  0.6676 &    0.05 &  0.8615 &  0.0045 &  0.4263 &  0.0347 &  0.3833 &  0.0712 &  0.5847 &      4.2 \\
MSCRED      &  0.5461 &  0.0602 &  0.6934 &  0.0056 &  0.3848 &  0.0263 &  0.0417 &       0 &  0.4165 &      8.8 \\
OmniAnomaly*           &   0.6451 &  0.1127 &  0.8294 &  0.0071 &  0.2232 &  0.0322 &  0.3528 &  0.0076 &       0.5126 & 7.8\\
\bottomrule
\end{tabular}
\end{table*}


\end{document}